%% file: example.tex
\documentclass{article}
\usepackage{amssymb}
\usepackage{graphicx}
\usepackage{xcolor}
\usepackage{siunitx}
\usepackage{algorithm}
\usepackage{algpseudocode}
\usepackage{subcaption}
\usepackage{amsmath}
\usepackage{tikz}
\usepackage{multicol}
\usepackage[bookmarks=true]{hyperref}
\usepackage{booktabs}
\usepackage[T1]{fontenc}
\usepackage{bm}

\definecolor{tab0}{RGB}{31,119,180}   % blue
\definecolor{tab1}{RGB}{148, 103, 189}   % orange
\definecolor{tab2}{RGB}{44,160,44}    % green
\definecolor{tab3}{RGB}{214,39,40}    % red

\colorlet{statecol}{tab3}    % red
\colorlet{futurecol}{tab2}   % green 
\colorlet{actioncol}{tab1}   % purple
\colorlet{paramcol}{tab0}    % blue

\usepackage[preprint]{corl_2025} % Uncomment for pre-prints (e.g., arxiv); This is like ``final'', but will remove the CORL footnote.

\include{definitions}

\title{Beyond Constant Parameters: \\
       Hyper Prediction Models and HyperMPC}

\author{
  Jan Węgrzynowski$^{1,2,3}$ \qquad
  Piotr Kicki$^{1,2,3}$ \qquad
  Grzegorz Czechmanowski$^{1,2,3}$ \\
  \textbf{Maciej Krupka}$^{1}$, \qquad   \textbf{Krzysztof Walas}$^{1,2}$ \\
  \\
  $^{1}$Institute of Robotics and Machine Intelligence, 
  Poznan University of Technology, Poland \\
  $^{2}$IDEAS Research Institute, Warsaw, Poland, \qquad
  $^{3}$IDEAS NCBR, Warsaw Poland \\
  \texttt{jan.wegrzynowski@put.poznan.pl} \\
}

\begin{document}
\maketitle

%===============================================================================
\vspace{-0.5cm}

\begin{abstract}
    Model Predictive Control (MPC) is among the most widely adopted and reliable methods for robot control, relying critically on an accurate dynamics model.
    However, existing dynamics models used in the gradient-based MPC are limited by computational complexity and state representation.
    To address this limitation, we propose the Hyper Prediction Model (HyperPM) - a novel approach in which we project the unmodeled dynamics onto a time-dependent dynamics model.
    This time-dependency is captured through time-varying model parameters, whose evolution over the MPC prediction horizon is learned using a neural network.  
    Such formulation preserves the computational efficiency and robustness of the base model while equipping it with the capacity to anticipate previously unmodeled phenomena.
    We evaluated the proposed approach on several challenging systems, including real-world F1TENTH autonomous racing, and demonstrated that it significantly reduces long-horizon prediction errors.
    Moreover, when integrated within the MPC framework (HyperMPC), our method consistently outperforms existing state-of-the-art techniques.
\end{abstract}


%===============================================================================

\section{Introduction}

One of the fundamental challenges in robotics is determining optimal actions to achieve specific objectives, a task generally referred to as control.  This challenge is particularly pronounced for systems that are underactuated or operate at the limits of their physical capabilities. Many of these complexities have been recently resolved with learning-based controllers~\cite{kober2013rlsurvey, jiang2020learning, brunke2022saferl, czechmanowski2025rlracing}.
However, if the model of the system is available, MPC has demonstrated outstanding performance across demanding tasks, such as agile drone flight~\cite{song_policy_2022, song_reaching_2023, krinner_mpcc_2024}, autonomous racing~\cite{kabzan_learning-based_2019-1, chrosniak_deep_2023}, and legged locomotion~\cite{neunert_whole-body_2018, grandia2019feedback_mpc}.

The effectiveness of any MPC-based approach fundamentally depends on the quality of the underlying dynamical system model, as its accuracy, robustness, and computational complexity critically influence the controller's performance.
In general, the accuracy of the dynamics model can be improved at the expense of the increased computational effort by the use of data-driven components, such as Gaussian Processes (GP)~\cite{hewing_cautious_2018, kabzan_learning-based_2019, torrente_data-driven_2021}, polynomials~\cite{kaufmann_champion-level_2023, krinner_mpcc_2024} or even small neural networks~\cite{spielberg_neural_2019, chee_knode-mpc_2022, salzmann_real-time_2023}. However, to preserve the sparsity of the optimization problem, each of these models predicts the system's evolution based on its current state and control.
This may not be enough to accurately capture the influence of unmodelled system states dynamics on the outputs we want to control, e.g., mass transfer during braking, tire deformation, suspension movement during aggressive car racing, or the position of the payload suspended from a drone by a rope.

\begin{figure}[t]
  \centering
  \includegraphics[width=\linewidth]{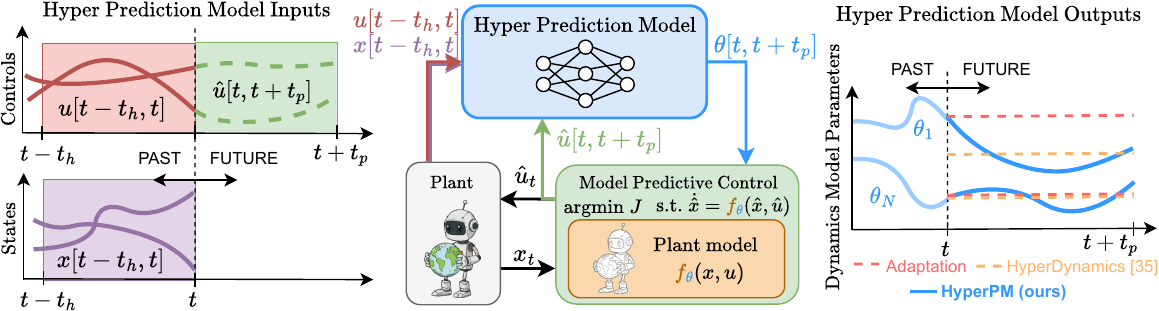}
  \vspace{-5mm}
  \caption{
    % HyperPM improves the prediction accuracy by infering dynamics model parameters $\textcolor{futurecol}{\hat{u}[\!t_c,\;t_c + t_p]}$ future trajectories. It utelizes the past observations to infer unobservable state of the sytem and expected control to anticipate unmodeled efects.
  % The Hyper Prediction Model, based on the history of observations and the expected control trajectory, predicts the future trajectories of the parameters  of the dynamics model, which are then used in MPC.\todo{citations in the right plot legend
  HyperPM leverages the recent state $\textcolor{actioncol}{x[t_c-t_h,t_c]}$ and control $\textcolor{statecol}{u[t_c-t_h,t_c]}$ history, together with the planned control sequence $\textcolor{futurecol}{\hat{u}[t_c,t_c+t_p]}$, to predict a trajectory of time-varying model parameters $\textcolor{paramcol}{\theta[t_c,t_c+t_p]}$. Injecting these predicted parameters into the nominal dynamics yields more accurate long-horizon state forecasts, allowing HyperMPC to proactively anticipate and compensate for previously unmodeled effects.
  }
  \label{fig:first_page}
  \vspace{-6mm}
\end{figure}

To address these limitations, we propose encoding the unmodeled components of the system dynamics as variations in the parameters of the existing model.
This allows us to increase the model's accuracy without incuring additional computational overhead during optimization.
However, the essence of MPC lies in leveraging the aforementioned models to predict the system's behavior over a finite but preferably long horizon. Therefore, predicting only the static parameters of the model may not be sufficient to work over extended prediction horizons. Thus, in this paper, we propose a Hyper Prediction Model (HyperPM) -- a framework that uses a neural network to predict the trajectories of the expected evolution of the model parameters, based on the history of observations and the planned trajectories of future control inputs (see Fig.~\ref{fig:first_page}). To the best of our knowledge, this is the first approach that goes beyond inferring the model based on its history, but also anticipates the expected evolution of the system. This allows us to obtain lightweight dynamics models with higher accuracy over the prediction horizon, enhancing MPC performance.

We evaluated the proposed approach on three challenging tasks: (i) swing-up of the simulated pendulum with backlash, (ii) trajectory tracking with a simulated drone with a payload suspended on a rope, and (iii) autonomous racing using a real F1TENTH car~\cite{f1tenth}. 
We show that the proposed HyperPM achieves over 33\%, {12\%}, and 62\% improvement in prediction accuracy in the pendulum, drone, and racing tasks, respectively, compared to an analytical model with constant parameters.
Moreover, we show that the use of HyperMPC results in a 300\% improvement over MPC with an analytical model in pendulum swing-up, {9\%} improvement in the case of the drone task, and almost 19\% improvement over MPC with a residual model in autonomous racing.

The contributions of this paper are the following:\\
 1) Hyper Prediction Model, a novel framework that predicts future trajectories of dynamics model parameters, based on the history of observations and planned actions, \\
 % to compensate for the effects of unmodeled system states., \\
 2) HyperMPC, which integrates HyperPM with optimization-based MPC to improve control performance without increasing the computational complexity of the dynamics model inside the optimization loop.

\vspace{-4pt}
\section{Related work}
\label{sec:citations}

\textbf{Model Predictive Control} is a widely adopted approach for controlling complex dynamical systems~\cite{schwenzer2021mpcsurvey}. In general, MPC algorithms can be categorized into two distinct groups: (i) sampling-based MPC and (ii) optimization-based MPC.
The sampling-based MPC approaches, such as MPPI~\cite{williams2016mppi_agresived_driving, williams2017mppi_information} or iCEM~\cite{pinneri2020icem}, offer great flexibility in terms of the dynamics models they use.
However, this flexibility comes at the cost of increased computational demand and typically the need to use powerful GPUs to enable parallel simulation of thousands of candidate control trajectories~\cite{williams2016mppi_agresived_driving, williams2017mppi_information, wang_pay_2023, williams_information_2017, xiao2024anycar}.
An alternative approach is to take advantage of numerical optimization instead~\cite{schwenzer2021mpcsurvey}. 
However, gradient-based MPC approaches are typically limited to the use of reduced-order analytical models of the controlled system~\cite{grandia2019feedback_mpc, hanover2022adaptivempc} or small function approximators \cite{kabzan_learning-based_2019-1, krinner_mpcc_2024, salzmann_learning_2024}.

In this paper, we focus on the key aspect of both aforementioned MPC approaches: developing models for accurate predictions over long horizons.
However, since the proposed method's benefits are most pronounced when using lightweight reduced-order analytical models, we integrate our proposed HyperPM into optimization-based MPC and call it HyperMPC.

%%%%%%%%%%%%%%%%%%%%%%%%%%%%%%%%%%%%%%%%%%%%%%%

\textbf{Dynamics models} play a key role in MPC, as they are expected to accurately predict the evolution of the system within a limited computational budget. 
% Their primary objective is to ensure the highest possible accuracy within a limited computational budget.
A conventional approach in MPC is to use reduced-order analytical dynamics models derived from first principles~\cite{verschueren2016racingmpc, liniger2015optimization, schwenzer2021mpcsurvey}. 
% These models typically trade off accuracy for computational efficiency, interpretability and robustness.
At the other end of the spectrum, there are neural-network based dynamics models~\cite{hansen2024tdmpc2, zhang2019solar, xiao2024anycar}.
% \todo{add some recent citatinions}.
% , which typically trade off computational efficiency and robustness for increased accuracy.
Somewhere between these two extremes, we find physics-based analytical models augmented with some general function approximators, such as Gaussian Processes \cite{hewing_cautious_2018, kabzan_learning-based_2019, torrente_data-driven_2021}, polynomial basis expansions \cite{krinner_mpcc_2024, kaufmann_champion-level_2023}, or neural networks \cite{spielberg_neural_2019, chee_knode-mpc_2022, salzmann_real-time_2023}. These hybrid approaches aim to balance the accuracy of the data-driven models with the computational efficiency and robustness of the analytical ones.

Nevertheless, the approximation capabilities of all these models are limited by the system's partial observability. This limitation may be overcome by including the history of observations~\cite{spielberg_neural_2019, xiao2024anycar}, but doing so is infeasible in the context of optimization-based MPC~\cite{chee_knode-mpc_2022, salzmann_real-time_2023}.
An intriguing approach to address this issue is to use hyper-neural networks~\cite{zhang_graph_2020, hegde_hyperppo_2024, xian_hyperdynamics_2021}, which can predict the parameters of other neural networks that act as the dynamics model~\cite{xian_hyperdynamics_2021}.
% Hyper-networks have shown promise in architecture search \cite{zhang_graph_2020}, reinforcement learning \cite{hegde_hyperppo_2024}, and dynamics modeling \cite{xian_hyperdynamics_2021}.
A similar concept was presented in~\cite{chrosniak_deep_2023}, where a neural network was used to predict certain parameters of the analytical model.
However, both of these approaches assume predicting static parameters across the prediction horizon.

In our work, we build upon the ideas of~\cite{xian_hyperdynamics_2021} and~\cite{chrosniak_deep_2023}, and exploit analytical models with parameters identified by a neural network, as in~\cite{chrosniak_deep_2023}. However, we demonstrate that one can improve the long-horizon prediction accuracy of a dynamics model by capturing the unmodelled dynamics through the expected evolution of model parameters over the prediction horizon. 

Although the proposed technique resembles online parameter adaptation or incremental model learning~\cite{triMAML2024, kabzan_learning-based_2019}, these existing methods assume a single, time-invariant model throughout the entire prediction horizon.
Our approach is orthogonal to these methods: it exploits knowledge of the planned control sequence to synthesise a time-varying dynamics model that is locally adapted to that sequence, thereby improving predictive accuracy. Crucially, nothing in the framework prevents online learning or weight adaptation $\Theta$, allowing further refinement while retaining its expressive power.

% While the proposed approach may seem similar to the online adaptation of the model parameters or online model learning\todo{add citations}, these approaches still assume the constant model along the prediction horizon. In fact, our approach is orthogonal to these and propose a way of exploiting the knowledge about the planned actions to infer a more accurate dynamics model around them.

%===============================================================================

% dron z payloadem na lince
% w tym przypadku na podstawie stanu drona nadal nie znamy pozycji i pręskości ładunky nie Markovian przez co
% co nie pozwala modelowi na ich kompensację ich efektu.

% można by to odzyskać z histori to ze względu na chęc użycia numerical solver based mpc nie możemy mieć modelu który bierze długa historię ze względu ograniczenia obliczeniowe, 

% ponadto planując ruch warto znac jak ewolucja nie modelowanych stanów nie tylko w aktualnej chwili ale też w całym horyzoncie predykcji będzie wpływać na model.

% Metoda ---------------------------------------------------------------------------

\vspace{-5pt}
\section{Proposed Method}
\label{sec:proposed_method}

\subsection{Considered problem}
In this paper, we address the problem of identifying a dynamics model $ \dxpred = f_\theta(\xpred, u)$, parametrized by $\theta$, which accurately predicts the evolution of the observable system state $x$ over the long prediction horizon. This can be formalized by
\begin{equation}
\label{eq:problem}
    \argmin_\theta \int_{\tc}^{\tc+\tp} \|\xt - \xpredt\|_2 dt \quad
    \text{s.t.} \quad \dxpred = f_\theta(\xpred, u),
\end{equation}
where $\tc$ represents the current time, $\tp$ the duration of the prediction horizon, e.g., 2 seconds, and $\xpred$ represents the predicted evolution of the system state. 
While this problem may be foundational to many downstream applications, we consider it within the context of Model Predictive Control, which exploits the long-horizon predictions to find an optimal control sequence.

To motivate our approach, consider a drone carrying a rope-suspended payload, where the observable state includes only the drone state variables - position, orientation, and linear and angular velocities.
Consequently, a model relying solely on $(x_t,u_t)$ will inevitably incur errors, because the influence of the swinging payload cannot be accurately captured without access to its unobserved state.
One potential remedy is to employ a model that conditions on a finite history of observations; however, such models are computationally prohibitive inside the optimisation loop of solver-based MPC.  
To effectively use MPC, one would require the payload’s influence not just at the current time step, but throughout the entire prediction horizon.
Consequently, any domain characterized by partially observed dynamics will encounter analogous prediction errors and degraded control performance when used within the MPC framework.

\newpage
\subsection{Hyper Prediction Model}

Continuing with the running example of a drone transporting a rope-suspended payload, one might initially treat the payload-induced forces as exogenous disturbances and entirely omit them from the system model.
Although these forces originate from the payload’s own swing dynamics and appear indirectly in the drone’s measured states. If the payload state were observable, its equations of motion—and hence its influence on the drone's airframe could, in principle, be derived and incorporated into the model.
Nevertheless, a model that relies only on the drone’s instantaneous state cannot capture those effects.
An approach is to compensate for unmodeled states by allowing the dynamics model parameters $\theta$ formulated in \eqref{eq:problem}, to vary in response to those effects.
However, most existing methods that address model inaccuracies, whether via adaptation laws~\cite{l1quad}, online parameter prediction~\cite{xian_hyperdynamics_2021, chrosniak_deep_2023}, or other forms of online learning~\cite{triMAML2024, kabzan_learning-based_2019}, assume that these parameters remain constant over the entire prediction horizon.
We challenge this assumption by instead inferring the entire future trajectory of unmodeled states over the prediction horizon, conditioning not only on past observations but also on the expected control sequence. The impact of these unmodeled states is then captured as a trajectory of time-varying model parameters, yielding the Hyper-Prediction Model ($\HyperPM$):
\begin{equation}
\begin{aligned}
    {
    % \theta[\tc, \tc + \tp] \;=\;
      % \HyperPM_{\Theta} \!\bigl(x[\tc - \thist, \tc],\, u[\tc-\thist,\tc], \, \hat{u}[\tc, \tc + \tp]\bigr),
      \textcolor{paramcol}{\theta[t_c,\;t_c + t_p]} \;=\;
  \HyperPM_{\Theta} \Bigl(
    \textcolor{statecol}{x[t_c - t_h,\;t_c]},\;
    \textcolor{actioncol}{u[t_c - t_h,\;t_c]},\;
    \textcolor{futurecol}{\hat{u}[t_c,\;t_c + t_p]} \Bigr) 
    }
\end{aligned}
\end{equation}
where $\HyperPM$ ingests the recent state history $x[\tc-\thist,\tc]$, the corresponding control history $u[\tc-\thist,\tc]$, and the planned control sequence $\hat{u}[\tc,\tc+\tp]$. Graphically presented in Figure \ref{fig:first_page}.
The $\HyperPM$ model (i) extracts a latent representation of the unobserved unmodeled state from this history, (ii) propagates that representation forward under the planned actuation, and (iii) expresses the resulting influence of unmodeled states as trajectory of time-varying parameters $\theta[t_c,\;t_c + t_p]$ across the prediction horizon.
Additional implementation details are provided in the Appendix \ref{apx:architecture}.

% Our implementation of the $\HyperPM$ first encodes past observations with a GRU; its hidden state is then passed, combined with future actions to a masked MLP that preserves causality with respect to the anticipated action sequence.
% Control inputs and predicted parameters are temporally compressed using a B-spline basis, and zero-mean Gaussian noise is injected into the down-sampled future actions to bolster robustness against action-prediction errors. 

It is important to emphasize that our approach goes beyond even oracle adaptation; as the experimental results show, even a perfect adaptive scheme is limited to simply propagating the obtained parameters as constant in time, for example, assuming that the payload forces that currently act on the drone frame will be the same in future. 

In summary, our method addresses scenarios where obtaining a fully Markovian representation of the system state is infeasible or computationally prohibitive. By approximating certain missing aspects of the dynamics within the model's parameter space, HyperPM improves the accuracy of long-horizon predictions by encoding unmodeled dynamics into trajectories of time-varying parameters.

\vspace{-5pt}
\subsection{Training procedure}
An essential component of the proposed solution is the training procedure for HyperPM. Because in our proposed solution, we predict the trajectory of model parameters, we cannot rely on the most common approach to training parameter prediction models, i.e., evaluating the accuracy on one-step predictions.
Instead, we propose to:
\begin{enumerate}
    \item predict the trajectory of model parameters $\theta[\tc, \tc+\tp]$, based on the history of states $\x[\tc - \thist, \tc]$, actions  $u[\tc - \thist, \tc]$  and future controls $u[\tc, \tc + \tp]$,
    \item roll out the time-varying model dynamics $f_{\theta_{t}}$ in discrete time steps defined by for exmple Runge Kutta 4-th order integration scheme $\xpred_{t + dt} = f_{\theta_{t}}^{\RK}(\xpred_t, u_t)$,
    \item compute the mean square prediction error between the obtained predicted trajectory of states $\xpred[\tc + dt, \tc + \tp]$, with the ones from the dataset $x[\tc + dt, \tc + \tp]$,
    \item compute the loss as a sum of the prediction error and regularization terms,
    \item back-propagate through time~\cite{werbos_backpropagation_1990} the gradients of the loss function w.r.t. HyperPM weights and biases $\Theta$.
\end{enumerate}
% This procedure is schematically depicted in Appendix. %Figure~\ref{fig:rollout_training_scheme}.
By applying this training scheme, we ensure that every trainable variable is directly optimized to improve long-horizon prediction performance, which is crucial for effective MPC. A detailed description of the training procedure is provided in the Appendix \ref{apx:architecture}.

% Moreover, the proposed approach has no restrictions regarding the dynamics model structure.

\subsection{HyperMPC}

The downstream application of HyperPM, which we focus on in this paper, is optimization-based Model Predictive Control~\cite{schwenzer2021mpcsurvey, acados}. In this framework, an optimization algorithm computes trajectories of states and control actions that minimize the objective function, while satisfying the constraints stemming from the robot's dynamics and imposed on its states and control signals. In order to close the feedback loop, only the first control action is applied to the plant, and the whole optimization process is repeated at the next control interval. 

The key aspect of the MPC paradigm is the use of a model capable of accurately predicting the evolution of the system over the optimization horizon. Typically, MPC uses a single constant model of the system~\cite{schwenzer2021mpcsurvey}. However, as shown in~\cite{xian_hyperdynamics_2021}, it is possible to infer a new model at each time step, so that it can capture local variations in the dynamics of the system. In our approach, called HyperMPC, we propose to go further and exploit HyperPM to predict the trajectory of model parameters over the MPC horizon and increase predictive accuracy not only locally but also in the expected near future.

% To do so we utilize the trajectory of planned actions from the previous iteration of MPC solution. We feed this trajectory of planned actions and history of observation to HyperMode to get the trajectory of predicted parameters witch are used in next optimiation sceem

% We then use the trajectory of predicted paramters from HyperModel to 

% We then use the trajectory of planed actions from the previous iteration of MPC solution, to infer the dynamic model parameters that are fed into MPC framework, next optimal control problem is being solve and the first action if forwarded to the control plant and the whole cycle continues. 

We use the planned control sequence generated by the previous MPC iteration and, together with the most recent observation window, pass it to $\HyperPM$ to obtain a horizon-long trajectory of dynamics-model parameters.
These predicted parameters are then injected into the nominal dynamics model and supplied to the MPC solver, which resolves the optimal-control problem for the current horizon.
Only the first action of the refreshed solution is applied to the plant; the state and planned-action buffers slide forward, and the entire cycle repeats at the next control interval.

Crucially, the scheme maintains the same computational complexity as the primary dynamics model within the optimization loop, adding only a single forward pass through a lightweight neural network performed before solving the optimal control problem.

\section{Experiments and results}

\begin{figure}[h]
    \centering
    \includegraphics[height=33mm]{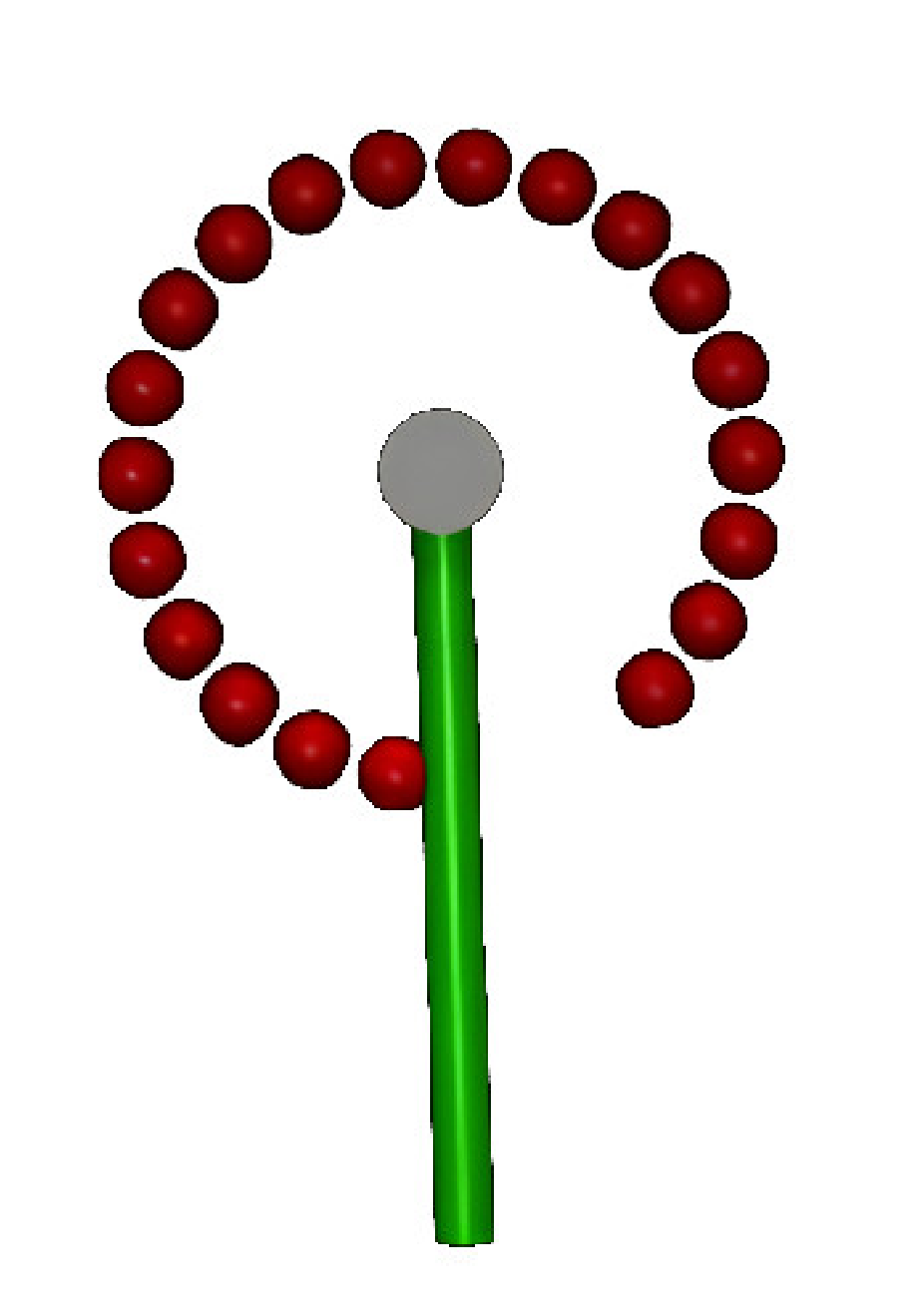}
    \includegraphics[height=33mm]{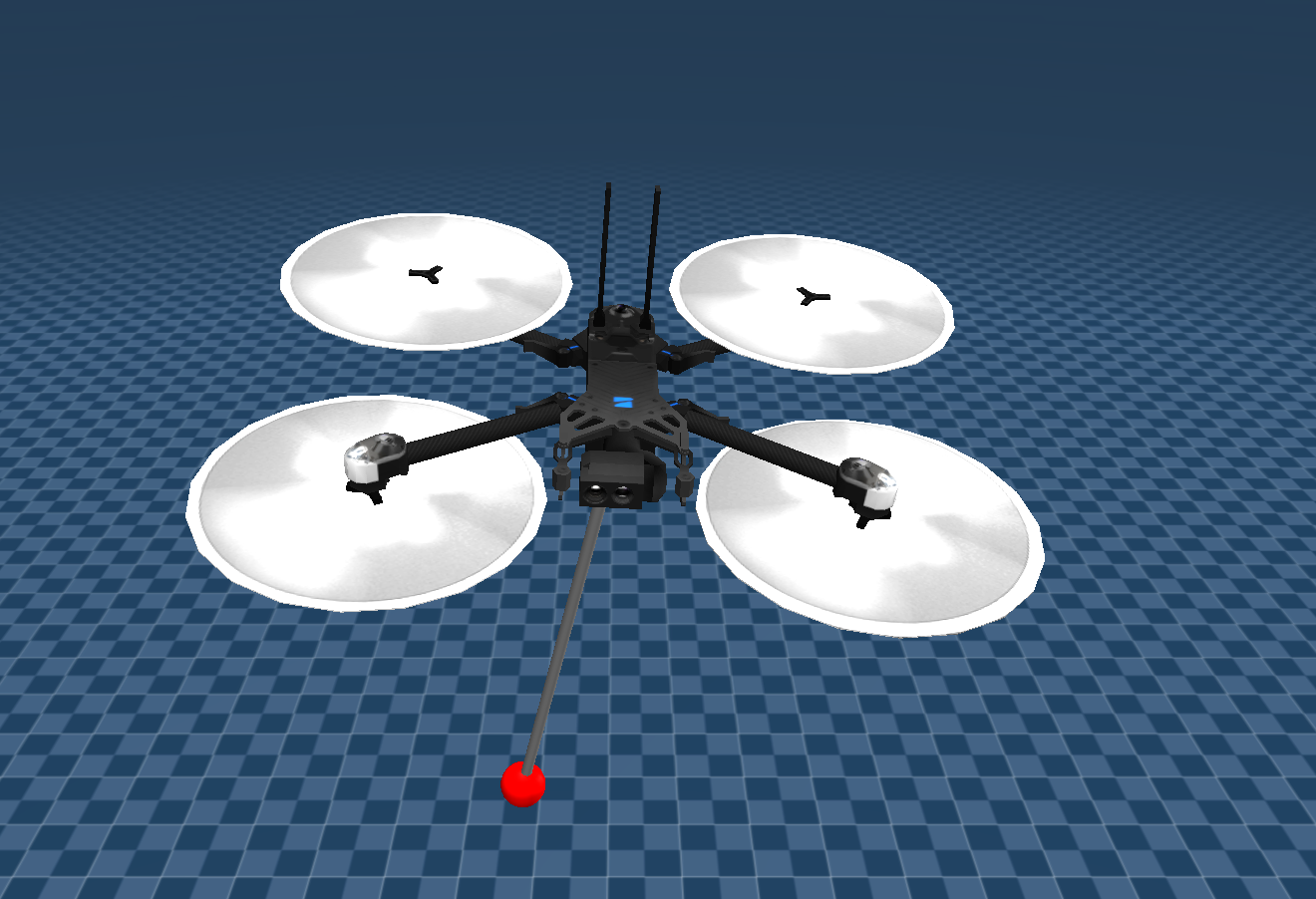}
    \includegraphics[height=33mm]{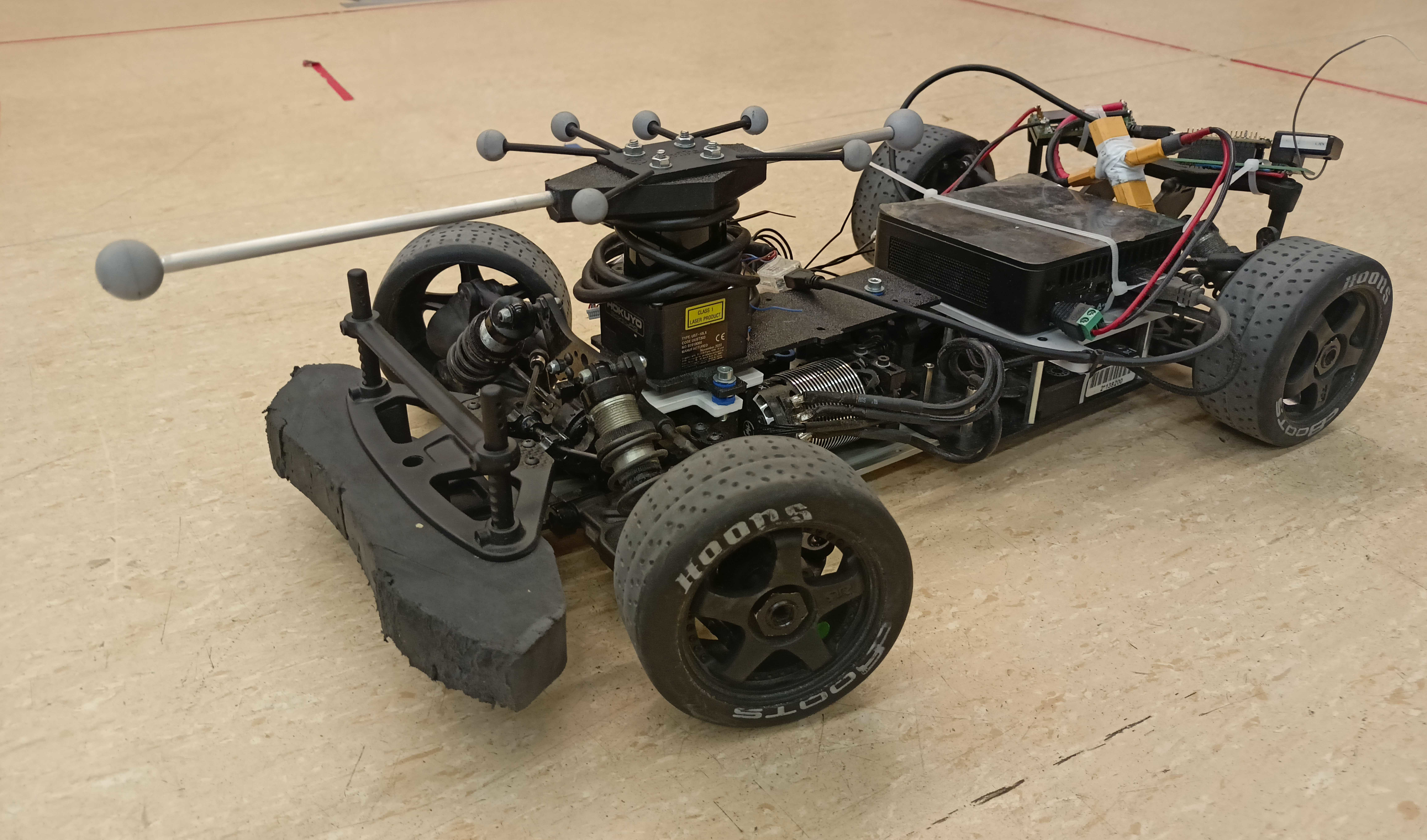}
    \caption{Experimental platforms used to validate the proposed method: (i) pendulum with backlash swing-up, (ii) drone with rope attached payload trajectory tracking, (iii) real-world autonomous F1TENTH~\cite{f1tenth} racing.}
    \label{fig:all_experiments_pic}
\end{figure}

To evaluate the effectiveness of the proposed HyperPM and HyperMPC, we performed experiments on a set of challenging control tasks presented in Fig.\ref{fig:all_experiments_pic}.
% : (i) swing-up of a simulated pendulum with backlash, (ii) drone with rope attached payload trajectory tracking and (iii) autonomous racing using a real-world F1TENTH car~\cite{f1tenth}.
% This section outlines the experimental setup, evaluation metrics, and results obtained. Across all scenarios, we evaluated prediction accuracy over long sequences and control performance in downstream tasks when integrated with an MPC framework.
We show that forecasting the trajectories of dynamic model parameters using $\HyperPM$ enhances the prediction accuracy and improves the control performance in MPC.
In each experiment, control performance is evaluated via the acumulation of MPC stage cost across episodes, highlighting differences between the modeling approaches.

\subsection{Baselines}
\label{sec:baselines}

To establish baselines for our method, we selected a group of approaches to model the dynamics of the considered systems:
    \textbf{$\consts$} -- A dynamics model with constant parameters, obtained using \underline{s}ingle-step prediction error minimization.
    \textbf{$\constl$} -- A dynamics model with constant parameters, optimized using \underline{l}ong sequences to minimize the error between simulated system states and ground-truth states across entire trajectories.
    \textbf{$\HDs$ }-- Referred to as HyperDynamics, inspired by \cite{chrosniak_deep_2023, xian_hyperdynamics_2021}. This is a dynamics model in which parameters are predicted by a neural network and held constant for the entire prediction horizon of the roll-out. It is trained using single-step prediction error minimization according to \cite{chrosniak_deep_2023, xian_hyperdynamics_2021}.
    \textbf{$\HDl$} -- The same architecture as $\HDs$, but trained using long sequences.
    \textbf{$\HyperPM$} -- Our proposed method, as described in Section \ref{sec:proposed_method}.

In addition, we tested extending the nominal dynamics of the system with a residual neural network~\cite{chee_knode-mpc_2022, salzmann_real-time_2023} $\dot{x} = f(x, u) + \NN(x, u)$ trained using long sequences: $res$. For drone experiments, the residual network predicts the residual forces acting on the drone frame.

To ensure a fair comparison with baselines, the training and architectural hyperparameters (e.g., neural network architecture, batch size, and learning rate) were optimized for each method by grid search, using the accuracy of the validation subset as the evaluation criterion.
Full details of the training setup including hyper-parameters, exact MPC formulation, and dataset descriptions are provided in the Appendix \ref{apx:all}.

\vspace{-2pt}
\subsection{Simulated pendulum with backlash}

The first illustrative example of a system where all previous methods are constrained by the assumption of constant model parameters across the prediction horizon is a pendulum with backlash. Backlash refers to a clearance-induced dead zone causing temporary torque loss during motion reversal. To effectively plan a swing-up, the controller needs to anticipate the effect, something that static parameter models inherently fail to capture.

% \subsubsection{Long-horizon prediction results}

We evaluated the predictive performance of the models using a dedicated test set comprising sequences of 250 samples at 100 Hz. The results of this comparison are presented in Table~\ref{tab:pendulum_combined_clean}.
All models, except $\text{HD}_s$ and our $\text{HyperPM}$ method, performed comparably well in the long-horizon prediction task. The $\text{HD}_s$ model, trained on short sequences, was unable to produce coherent long-sequence predictions due to overfitting to single-step predictions.
In contrast, only our methods demonstrated significantly better performance, as they could recover the underlying state of the backlash and effectively compensate for it by leveraging time-varying dynamics model parameters, demonstrating the strength of forecasting parameter trajectories over the prediction horizon.

\begin{table}[h]
\centering
\vspace{-3mm}
\caption{Prediction and MPC performance for a pendulum with backlash}

\scalebox{0.8}{
\begin{tabular}{lrr|ccc}
\toprule
\multicolumn{3}{c|}{\textbf{Long-horizon prediction}} & \multicolumn{3}{c}{\textbf{MPC performance in the swing-up task}} \\
Model &
\begin{tabular}{@{}c@{}}Error ($\downarrow$)\end{tabular} & 
\begin{tabular}{@{}c@{}}Improvement [$\%$] ($\uparrow$)\end{tabular} &
\begin{tabular}{@{}c@{}}Cost (mean\,±\,std) ($\downarrow$)\end{tabular} & 
\begin{tabular}{@{}c@{}}Success Rate  [$\%$] ($\uparrow$)\end{tabular} & 
\begin{tabular}{@{}c@{}}Solver \\ Failure Rate [\%] ($\downarrow$)\end{tabular} \\
\midrule
$\consts$           & 0.670 & 0.00    & 105.57 ± 10.96 & 0.0 & 5 \\
$\constl$           & 0.588 & 12.23   & 106.17 ± 1.92  & 0.0 & 15 \\
$\HDs$              & 28.766 & -4195.94 & 115.54 ± 10.08 & 0.0 & 45 \\
$\HDl$              & 0.626 & 6.51    & 85.92 ± 0.54  & 0.0 & 95 \\
oracle adaptation   & ---   & ---     & 94.10 ± 1.24  & 0.0 & 0 \\
$\res$              & 0.618 & 7.77    & ---            & --- & --- \\
$\HyperPM$ (ours)   & \textbf{0.447} & \textbf{33.19} & 34.98 ± 7.21  & \textbf{100} & \textbf{0} \\
% $\HyperPMnp$ (ours) & 0.475 & 29.11   & \textbf{33.27 ± 9.34} & \textbf{100} & \textbf{0} \\
\bottomrule
\end{tabular}
}
\normalsize
\label{tab:pendulum_combined_clean}

\vspace{-8pt}
\end{table}

% \subsubsection{MPC results}

We tested the different dynamics models within an MPC framework to evaluate their impact on a downstream control task.
We measured the total cost of each episode, the success rate (pendulum stabilized in upright position), 
% stage cost less than 0.01 at the end of the episode
and the failure rate of the MPC numerical solver.
The results obtained are presented in Table~\ref{tab:pendulum_combined_clean}, highlighting the advantages of time-varying parameter models in terms of both control performance and solver reliability.
We omitted the reimplementation of the $\res$ model for MPC as it does not show any improvement over the $\constl$ model in the prediction task.
To further illustrate the advantages of our method, we implemented oracle adaptation on top of a model that was trained on a dataset without backlash. Based on the simulator's internal state, we change the gear ratio parameter to 0 if the lever is not in contact with the gears.
None of the baselines handled the unmodeled dynamics or successfully swung the pendulum.
Even the oracle adaptation failed, as the controller couldn't anticipate unmodeled effects over the MPC horizon.
In contrast, our methods, which incorporate time-varying parameter trajectories, successfully produced an accurate model and stabilized the pendulum in the upright position across all test cases. Moreover, our solutions were also the most stable solutions that did not cause any solver failures, which were common for the baseline models, especially HD.

\subsection{Simulated drone with rope suspended payload}

Drone trajectory tracking is a fundamental component of an aerial vehicle's autonomy stack and presents a particularly challenging modeling problem, especially when the drone is tasked with transporting a rope-suspended payload. In our setup, the rope was attached to the center of mass of the drone frame, allowing us to model the behavior using only forces. It is important to note that the state of the system does not include the position or velocity of the payload.
This partial observability introduces complex, time-varying disturbances not evident from the drone’s state alone, making accurate long-horizon prediction and control challenging for traditional MPC approaches.

In comparison, we used two dynamics models, nominal~\cite{delayMPCdrone2020}, and a residual model in which a neural network predicts drone frame residual forces. Parameters that can be changed by $\HDl$ or $\HyperPM$ model are the additional forces acting on a drone's frame. Additionally, we include an oracle-adaptation baseline that retrieves the exact payload-induced forces from the simulator and assumes that these forces remain constant throughout the prediction horizon. This represents an idealized scenario with access to perfect information, albeit with the same limitations of time-invariant parameterization.
We omitted the models trained using single-step prediction as they proved unstable inside MPC loop and failed to deliver reliable control performance.

\begin{minipage}[t]{0.47\textwidth}
    
    \vspace{0.50cm} % Forces top alignment
    \centering
    \scalebox{0.7}{
    \begin{tabular}{lrr}
    \toprule
        Nominal Model & Error (mean\,±\,std) ($\downarrow$) & Improvement [\%] ($\uparrow$) \\
    \midrule
        $\constl$           & $0.0655 \pm 0.132$ &   --- \\
        $\HDl$              & $0.0252 \pm 0.083$ &  61.52  \\
        $\HyperPM$ (ours)   & $\textbf{0.0176} \pm 0.076$ &  73.12 \\
    \toprule
        Residual Model & Error (mean\,±\,std) ($\downarrow$) & Improvement [\%] ($\uparrow$) \\
    \midrule   
        $\constl$           & $0.0186 \pm 0.058$ &   --- \\
        $\HDl$              & $0.0184 \pm 0.057$ &   1.08 \\
        $\HyperPM$ (ours)   & $\textbf{0.0164} \pm 0.060$ &   11.82 \\
    \bottomrule
    \end{tabular}
    }
    \vspace{0.75cm}
    \captionof{table}{Long-horizon prediction error for drone with payload, improvement is calculated wrt. $\constl$ model with appropriate dynamics model.}
    \label{tab:drone_test_set}
    
    \end{minipage}%
    \hfill
\begin{minipage}[t]{0.51\textwidth}

    \vspace{-4pt}
    \includegraphics[width=\linewidth]{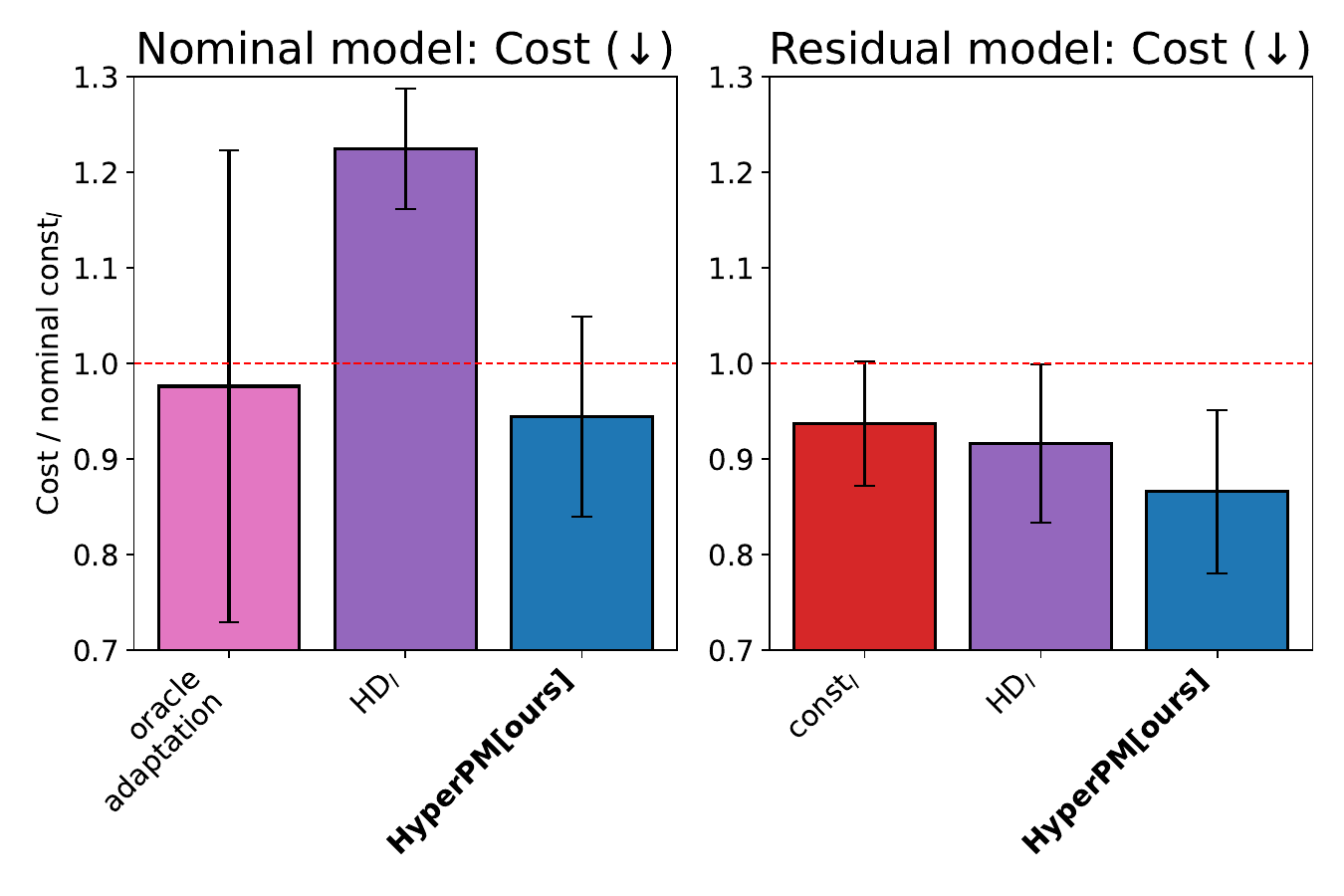}
    \captionof{figure}{Comparison of modeling approaches drone on trajectories tracking task relative to nominal $\constl$.}
    \label{fig:res_drone_mpc}
\end{minipage}

The reported results are based on a setup that employs a 0.5kg payload suspended from a 1m rope attached to a quadrotor of mass 1.325kg; additional rope‑length ablation studies are presented in the Appendix C.5.
In the long‑horizon prediction task (1s horizon, 100‑sample episodes), both HDL and HyperPM substantially improve upon the nominal model. However, for residual dynamic modeling, only HyperPM delivers a clear gain, exceeding all other methods by more than 10\%, underscoring its ability to anticipate complex, time-varying unmodeled effects. In the zero-length rope configuration (that is, the payload rigidly attached to the vehicle), we did not observe any measurable improvement, suggesting that HyperPM’s benefit is primarily due to its ability to compensate for previously unmodeled dynamics associated with payload-swing states.
For the downstream task evaluation, we used the test split of the generated dataset and formulated a trajectory‑tracking problem via MPC following the formulation in \cite{delayMPCdrone2020}. As presented in Figure. \ref{fig:res_drone_mpc}, our approach surpasses both the oracle‑adaptation baseline and $\HDl$ in both the nominal and residual dynamics model settings. These results further confirm the advantage of modeling time-varying parameter trajectories in capturing complex, unobserved dynamics and translating that capability into superior control performance.

\subsection{F1TENTH racing} 

Autonomous racing represents a unique and highly demanding testbed for evaluating and advancing control algorithms.
Unlike conventional autonomous driving scenarios, racing pushes vehicles to operate at the edge of their dynamic limits, requiring algorithms to manage highly nonlinear dynamics and unrecoverable consequences of actions. The subsequent experiments use models trained on a 36-minute dataset collected using a real F1TENTH vehicle, with expert drivers manually navigating a variety of racetracks to span the entire operational envelope of the vehicle.
Our model-based controller objective is to maximize track progress while keeping the vehicle within track boundaries.

% \subsubsection{Long-horizon prediction results}

\begin{figure}[h]
\centering

\begin{minipage}[t]{0.47\textwidth}

    \vspace{0pt}
    % \vspace{0.001cm} % Forces top alignment
    % \centering
    \captionof{table}{Long-horizon prediction error for F1TENTH car}
    \label{tab:car_test_set}

    \vspace{-8pt}

    \scalebox{0.7}{
    \begin{tabular}{@{}lcc@{}}
    \toprule
    Model & Error (mean\,±\,std) ($\downarrow$) & Improvement [\%] ($\uparrow$) \\
    \midrule
    $\consts$     & 0.0240 $\pm$ 0.0581  & 0.00     \\ 
    $\constl$     & 0.0177 $\pm$ 0.0290  & 26.25    \\ 
    $\HDs$        & 0.0501 $\pm$ 0.1880  & -108.75  \\ 
    $\HDl$        & 0.0137 $\pm$ 0.0239  & 42.92    \\ 
    $\res$        & \textbf{0.0091} $\pm$ 0.0138 & \textbf{62.08} \\ 
    % $\HyperPMnp$ (ours) & 0.0147 $\pm$ 0.0225 & 38.75    \\ 
    $\HyperPM$ (ours)   & \underline{0.0123} $\pm$ 0.0192 & \underline{48.75} \\
    \bottomrule
    \end{tabular}
    }

    \captionof{table}{Comparison of MPC solve time, and parameter inference time on mobile AMD Ryzen 5 4600HS CPU.}
    \label{tab:car_solve_time}

    \scalebox{0.7}{
    \begin{tabular}{@{}lcc@{}}
    \toprule
    Model & Solve time [ms] ($\downarrow$) & Inference time [ms] ($\downarrow$) \\
      & (mean\,±\,std) & (mean\,±\,std) \\
    \midrule
    $\consts$     & \textbf{10.76 }$\pm$ 2.24  & --     \\ 
    $\constl$     & \textbf{10.53} $\pm$ 1.97  & --    \\ 
    $\HDs$        & 13.42 $\pm$ 3.53  & 1.56 $\pm$ 0.13  \\ 
    $\HDl$        & \textbf{10.65} $\pm$ 2.05  & 1.48 $\pm$ 0.14  \\ 
    $\res$        & 23.73 $\pm$ 4.76  & -- \\ 
    % $\HyperPMnp$ (ours) & 0.0147 $\pm$ 0.0225 & 38.75    \\ 
    $\HyperPM$ (ours)   & \textbf{10.61 }$\pm$ 1.76 & 1.85 $\pm$ 0.12  \\
    \bottomrule
    \end{tabular}
    }

\end{minipage}%
\hfill
\begin{minipage}[t]{0.51\textwidth}

    \vspace{0pt}
    \includegraphics[width=\linewidth]{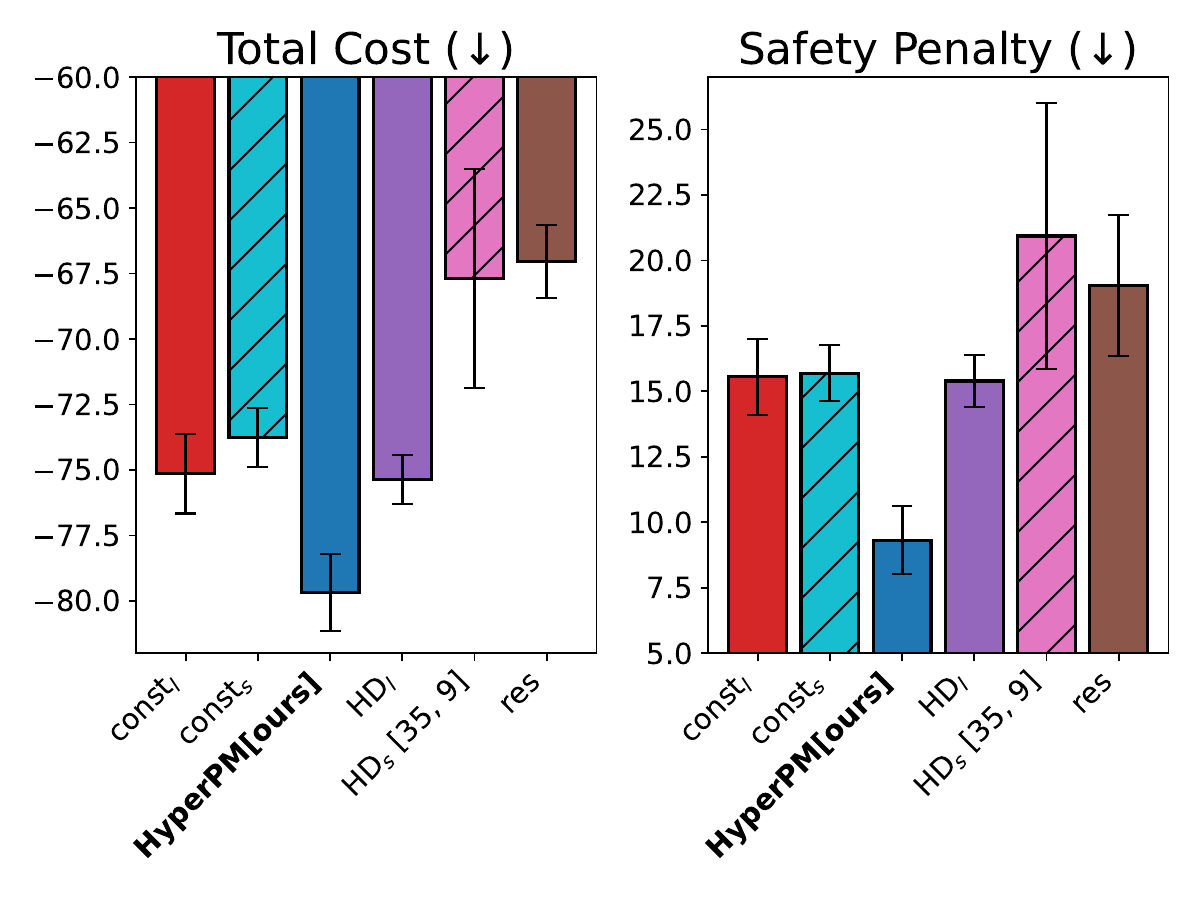}
    \captionof{figure}{Comparison of modeling approaches in the real-world F1TENTH racing task. }
    \label{fig:res_car_mpc}
    
\end{minipage}

\end{figure}

In our experimental analysis, we first evaluate the prediction performance of the models on long-horizon sequences drawn from the manual driving test set.
In Table \ref{tab:car_test_set}, we present the prediction errors for all evaluated modeling approaches and refer them to the most classical one -- $\consts$. The best-performing model is the residual model, which utilizes the ability to model residual errors of the analytical model with a neural network. However, this approach takes advantage of significantly more parameters than the remaining methods. Among the equally sized models, the proposed HyperPM yields the lowest error, improving 48.75\% over the model with constant parameters.
% Moreover, one can see that the utilization of the future controls is an important feature of the proposed method, as without it, it performs similarly to HyperDynamics trained on long sequences.
% Additionally, similarly as for the pendulum, one can observe that all models trained with long sequences obtain a lower error than their single-step counterparts.

% \subsubsection{MPC results}

We also evaluated the considered modeling approaches in the downstream task, i.e., real-world F1TENTH racing.
We compared them in 30-second runs using two criteria: (i) total cost $\ell(x, u)$ and (ii) safety penalty $\ell_s(x, u)$, which mainly include track bounds violation soft constraint, both of them accumulated over the whole run.
In Figure~\ref{fig:res_car_mpc}, we report these metrics averaged over five runs for each method.
HyperPM outperforms all methods, including the residual network with previously lowest prediction errors.
We attribute this performance gap to the lack of generalization ability of a residual neural network to a different distribution of inputs w.r.t. the manual driving dataset. %, and second, difficulty of optimizing the trajectory with a much more complex model inside.
Regarding computational cost, Table \ref{tab:car_solve_time} shows that feeding a parameter trajectory to the MPC leaves the solver’s runtime essentially unchanged.

%===============================================================================

\vspace{-5pt}
\section{Conclusion}
\label{sec:conclusion}

% In this paper, we introduced the Hyper Prediction Model (HyperPM) and its integration into the HyperMPC framework, aimed at improving the predictive accuracy and control performance of systems with complex dynamics. By forecasting time-varying parameters, HyperPM balances efficiency and accuracy, overcoming the limits of traditional and neural methods. The experimental evaluations demonstrated that HyperPM significantly improves the accuracy of the long-term prediction, with reductions of more than 35\% in prediction errors compared to the models with constant parameters. Moreover, the HyperMPC framework delivered superior performance in downstream control tasks, outperforming state-of-the-art methods by 76\% in the pendulum swing-up tasks, 9\% in the drone with payload trajectory tracking, and achieving an 8\% improvement in F1TENTH autonomous racing.
% Our results highlight the importance of incorporating anticipated control trajectories into the prediction of model parameters, enabling more accurate and robust dynamics modeling.

In this paper, we introduced the Hyper Prediction Model (HyperPM) and its integration into the HyperMPC framework to enhance the predictive accuracy and control performance of systems with complex dynamics. By dynamically forecasting time-varying parameters of a model, HyperPM bridges the gap between computational efficiency and accuracy, addressing the limitations of both traditional and neural network-based approaches. The experimental evaluations demonstrated that HyperPM significantly improves the accuracy of the long-term prediction, with reductions of more than 35\% in the prediction errors compared to the models with constant parameters. Moreover, the HyperMPC framework delivered superior performance in downstream control tasks, outperforming state-of-the-art methods by 76\% in pendulum swing-up tasks, 9\% drone with payload trajectory tracking, and achieving an 8\% improvement in F1TENTH autonomous racing.
Our results highlight the importance of incorporating anticipated actions into the prediction of model parameter trajectories, allowing MPC to proactively anticipate and compensate for previously unmodeled effects.

% Our results show that using anticipated control trajectories to predict dynamic model parameters trajectories is key to accurately modeling system dynamics.
% This lets HyperPM better capture future behavior, especially with unmodeled dynamics, improving MPC performance.
% Additionally, the interpretability of our framework provides insights into unmodeled phenomena, providing a pathway for enhancing physical models.

% Finally, the proposed training procedure, which leverages predictions over long sequences and back-propagation through time, not only enables training HyperPM but also significantly improves the performance of the State-of-the-Art dynamics modeling.

%===============================================================================

\clearpage

\section{Limitations}
The main limitation of the proposed approach is the need to have access to relatively accurate predictions of future control inputs. During training, the dataset provides exact future actions, whereas during deployment, we rely on expected action trajectories, typically generated by the controller or planner.
These expected trajectories are inherently imperfect in representing true future control sequences, especially in long-horizon predictions, due to the closed-loop nature of control and the system's sensitivity to disturbances and modeling errors. To mitigate these issues, we proposed to randomize future control sequences using Gaussian noise during training. An open problem is how to measure and take into account the plausible distribution and uncertainty of the subsequent controls and their impact on the change of the dynamics model.

Another important limitation of the presented approach is the inevitable policy mismatch: the dataset was generated by one policy, whereas the deployed controller operates with the dynamics model learned from that data. Consequently, the state-action distribution encountered online will drift away from the training distribution, risking overfitting to the original data and degraded closed-loop performance. We believe that this issue can be mitigated by online learning that continuously adapts the $\HyperPM$ model to the trajectories produced by its own policy.

%  is ievitable distribution shift from change of used policy to collect dataset that could potentialy lead to overfeting on dataset and performing poorly in control task. Alhtow this could be negated by use od incremental model learning of our proposed aproach.

% Another important limitation of the presented approach is that it is trained to minimize the prediction error with constant weights for all state dimensions, which is probably not an optimal choice for the downstream task~\cite{katayama, wegrzynowski2024iros}.

% Finally, in the considered approach we do not utilize the predictions made in the previous time-step, which may result in sudden changes in the trajectories of the dynamics model parameters between consecutive MPC calls, potentially causing optimization issues.

% The acknowledgments are automatically included only in the final and preprint versions of the paper.
% \acknowledgments{If a paper is accepted, the final camera-ready version will (and probably should) include acknowledgments. All acknowledgments go at the end of the paper, including thanks to reviewers who gave useful comments, to colleagues who contributed to the ideas, and to funding agencies and corporate sponsors that provided financial support.}

%===============================================================================

\bibliography{manual_references}  % .bib

\newpage

\section*{Appendix}
\label{apx:all}

\subsection{HyperPM and HyperMPC implementation}
\label{apx:architecture}

\subsubsection{HyperPM Architecture}

\begin{figure}[h]
    \centering
    \includegraphics[width=\linewidth]{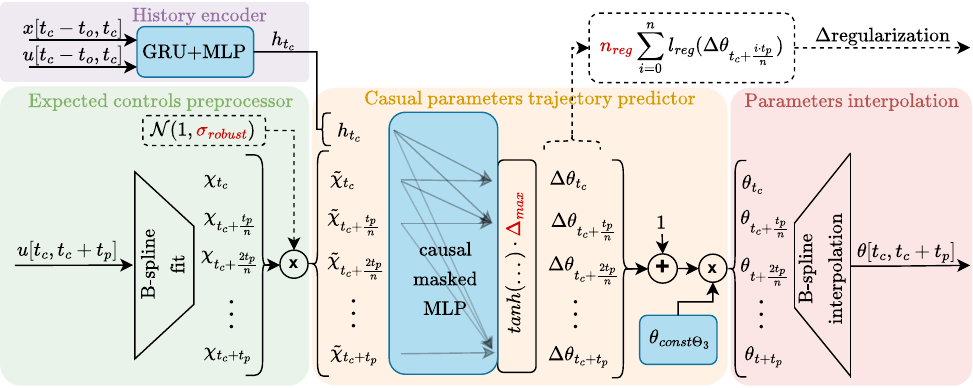}
    % \vspace{-4mm}
    \caption{The architecture of Hyper Prediction Model with trainable parts marked in blue and hyper-parameters marked in red.
    The central part of the data processing is done in the space of the B-spline control points. With dashed line we denoted the elements that are active only in the training phase.}
    \label{fig:hyper_model_architecture}
\end{figure}

In general, the HyperPM can be implemented with any machine learning model that predicts the expected trajectory of model parameters based on the history of states and actions, and the expected trajectory of actions.
In this paper, we utilize an architecture that is based on neural networks and we present its scheme in Figure~\ref{fig:hyper_model_architecture}.
The proposed architecture consists of four main components: (i) history encoder, (ii) expected controls preprocessor, (iii) casual parameters predictor, and (iv) parameters interpolation.

The main goal of the history encoder is to recover the underlying latent state of the system based on the history of states and actions. We examined multiple time-series encoders, including MLP, LSTM~\cite{hochreiter_long_1997}, and TCNN~\cite{bai_empirical_2018}, but finally we selected a recurrent neural network with GRU~\cite{chung_empirical_2014} for its superior performance.
The GRU's final hidden state is upscaled via a linear layer and is denoted as $h_{t_c}$ in Figure~\ref{fig:hyper_model_architecture}.

The goal of expected controls preprocessor is to generate a compact representation of the trajectory of future controls $u[\tc, \tc+\tp]$. Inspired by \cite{kicki_fast_2024, fey_splinecnn_2018}, we represent this sequence using control points of a B-spline curve fitted to the input data.
Thanks to the curve fitting procedure, we can freely change the sampling time of the input controls without retraining the HyperPM. Moreover, it reduces the number of temporal samples that need to be processed (resulting in a smaller network), and makes the network less sensitive to individual actions and more focused on the general trend. 

Before introducing the the second part of the control preprocessor, let us note that the inputs to the HyperPM are very different in their nature. Indeed, history of states and actions is known exactly, modulo measurement noise. Instead, the expected trajectory of future actions is much more undetermined.
During training, the dataset provides us with the exact future actions, however, during deployment we must rely on expected action trajectories, typically generated by the controller or planner.
Therefore, to account for the potential distributional shift and to make the HyperPM aware that the future controls are not as certain as the history of observations, we inject a noise to the representation of future actions.
Specifically, during training, we multiply the B-spline control points by the noise drawn from a normal distribution with a mean of 1 and a standard deviation of $\sigma_{\text{robust}}$.
We discovered that training with perturbed B-spline control points makes the network more robust and reduces overfitting (see Section~\ref{sec:sigma_ablation}).

% This design choice offers several benefits: it allows us to change the MPC's sampling time without retraining the neural network, reduces the number of temporal samples that need to be processed (resulting in a smaller network), and makes the network less sensitive to individual actions, which may be noisy.
% The second input to $\HyperPM$ is a sequence of future actions $u[\tc, \tc+\tp]$. 

% During training, the dataset provides exact future actions, whereas during deployment, we rely on expected action trajectories, typically generated by the controller or planner.
% These expected trajectories are inherently imperfect in representing true future control sequences, especially in long-horizon predictions.
% To address this, we make the neural network resilient to minor deviations in planned controls.
% Specifically, during training, we multiply the B-spline control points by the noise drawn from a normal distribution with a mean of 1 and a standard deviation of $\sigma_{\text{robust}}$.
% We discovered that training with perturbed B-spline control points not only makes the network more robust but also helps regularize it and reduce overfitting.

The following processing step in the HyperPM architecture is to generate changes in the parameters of the dynamics model over time, based on the representation of the state history $h_t$ and expected future controls. 
We do so using a neural network that consists of two MLP layers.
We enforce causality to prevent the model from learning spurious correlations or dataset-specific artifacts, such as driving style.
This is achieved by masking the weights that connect future actions to the past parameters, i.e., setting them to zero.
This ensures that the parameter trajectory generation remains causally dependent on the provided action sequences.

The representation of the changes of the parameters generated by the casual MLP is interpreted using two factors: scale $\Delta_{\text{max}}$ and bias $\theta_{\text{const}}$.
The scale $\Delta_{\text{max}}$ determines the maximum deviation from the nominal values of the model parameters $\theta_{\text{const}}$. For example, if $\Delta_{\text{max}} = 0.5$, the final parameter trajectory will be bounded between $50\%$ and $150\%$ of $\theta_{\text{const}}$.
The bias $\theta_{\text{const}}$ is the parameters vector that minimizes the average prediction error over the entire dataset.

Finally, the obtained values of the dynamics parameters are interpolated using B-splines and constitute the prediction of the evolution of the model parameters $\theta[\tc, \tc+\tp]$.
This design also enables one to extend the prediction horizon beyond the duration used during the training.
By replacing $\Delta \theta_{t + tp}$ with zero, the trajectory smoothly transitions to a constant parameter value.
The interpolated parameter vector can then be extended with the nominal parameters $\theta_{\text{const}}$ to achieve the desired prediction horizon.

Last but not least, we found it useful to regularize during training the changes in the parameters predicted by our solution using the L1 norm. This approach encourages HyperPM to predict changes in the parameters only if necessary and prefers to stay close to the nominal parameter values, as long as it does not affect the prediction accuracy too much. The regularization strength is controlled by $n_{reg}$.

\subsubsection{Training procedure}
An essential part of the proposed solution is the procedure that was used to train HyperPM. Because in our proposed solution, we predict the trajectory of model parameters, we cannot utilize the most common approach to training parameter prediction models, i.e., evaluating the accuracy on one-step predictions. Instead, we propose to:

\begin{enumerate}
    \item predict the trajectory of model parameters $\theta[\tc, \tc+\tp]$, based on the history of states $\x[\tc - \thist, \tc]$ and controls $u[\tc - \thist, \tc]$, and expected future controls $u[\tc, \tc + \tp]$,
    \item discretize $\theta[\tc, \tc+\tp]$ in subsequent time steps $\theta_{\tc}, \theta_{\tc+dt}, \ldots, \theta_{\tc + \tp - dt}$,
    \item roll out the time-varying model dynamics $f_{\theta_{t}}$ in discrete time steps using 4-th order Runge Kutta integration scheme $\xpred_{t + dt} = f_{\theta_{t}}^{\RK}(\xpred_t, u_t)$,
    \item compute the discretized prediction error~(see eqn. 1 in the main paper) between the obtained predicted trajectory of states $\xpred[\tc + dt, \tc + \tp]$ and the ones from the dataset $x[\tc + dt, \tc + \tp]$,
    \item compute the loss as a sum of the prediction error and $\Delta$regularization term introduced in Section~\ref{apx:architecture},
    \item back-propagate through time~\cite{werbos_backpropagation_1990} the gradients of the loss function w.r.t. HyperPM weights.
\end{enumerate}

\begin{figure}[h]
  \centering
  \includegraphics[width=\linewidth]{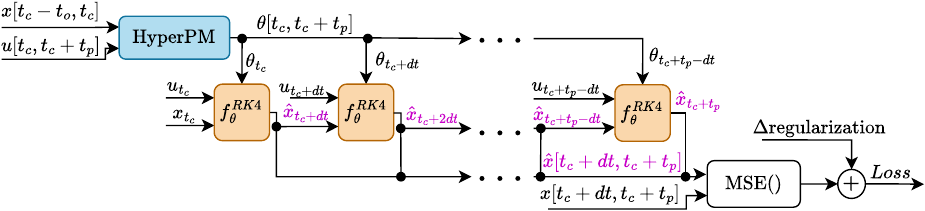}
  \vspace{-3mm}
  \caption{The Hyper Prediction Model is trained to predict trajectories of parameters of the dynamic model that minimize, over the prediction horizon, the mean squared error between simulated system states (pink) and the actual ones.}
  % The interior of HyperPM is illustrated in Fig. \ref{fig:hyper_model_architecture}.}
  \label{fig:rollout_training_scheme}
  \vspace{-2mm}
\end{figure}

This procedure is schematically depicted in Figure~\ref{fig:rollout_training_scheme}.
By applying this training scheme, we ensure that every trainable variable is optimized directly to improve long-horizon prediction, which is crucial for MPC. Moreover, the proposed approach has no restrictions regarding the dynamics model structure. Thus, we use the same approach to train all baseline approaches and show that optimization w.r.t. long-term loss and back-propagation through time is beneficial in terms of prediction accuracy and sometimes even necessary to obtain stable predictions in longer horizons.

\subsubsection{HyperMPC}

The downstream application of HyperPM, which we focus on in this paper, is optimization-based Model Predictive Control~\cite{schwenzer2021mpcsurvey, acados}. In this framework, an optimization algorithm computes the trajectory of states, and control actions, which minimize the objective function, while satisfying the constraints stemming from the robot dynamics and imposed on its states and control signals. In order to close the feedback loop, only the first action is forwarded to the plant, and the whole process is repeated at the next control interval. 

The key aspect of the MPC paradigm is a model that is able to accurately predict the evolution of the system over the optimization horizon. Typically, MPC utilizes a single constant model of the system~\cite{schwenzer2021mpcsurvey}. However, as shown in~\cite{xian_hyperdynamics_2021}, a new model can be inferred at each time step, so that it can capture local changes in the system's dynamics. In our approach, called HyperMPC, we propose to go beyond this and exploit HyperPM to predict the trajectory of model parameters over the MPC horizon and increase its accuracy not only locally but also in the expected near future.
In this way, not only MPC gets a better prediction model, but also HyperPM may improve the accuracy of its predictions by exploiting the knowledge of the actions planned by the MPC in the previous iteration.
The pseudocode of the HyperMPC algorithm is presented in Algorithm~\ref{alg:hypermpc}.

\begin{algorithm}[t]
\caption{HyperMPC Framework}
\label{alg:hypermpc}

\algrenewcommand\algorithmicrequire{\textbf{Input:}}
\algrenewcommand\algorithmicensure{\textbf{Output:}}

\begin{algorithmic}[1]
    
    \Require 
    \Statex Dynamics model: $f_\theta(x, u)$
    \Statex Objective function $\ell$ and the terminal objective $\ell_f$
    \Statex History of system states: $\mathcal{H} = \{\x_t\}_{t=\tc-\thist}^{\tc - dt}$
    \Statex Prediction horizon: $\tp$
    \Statex Current state: $\x_{\tc}$
    \Statex Expected controls: $\mathcal{C} = \{\hat{u}_t\}_{t=\tc}^{\tc+\tp-dt}$
    \Statex HyperPM: $\HyperPM(\mathcal{H}, \mathcal{C}) \rightarrow \{\theta_t\}_{t=\tc}^{\tc+\tp-dt}$
    
    % \Ensure Optimal control sequence: $\{u_k^*\}_{k=0}^{N-1}$
    \Ensure Control $u_{\tc}$
    
    \State \textbf{Predict Trajectory of Model Parameters}
    \[
    \{\theta_t\}_{t=\tc}^{\tc+\tp-dt} \gets \HyperPM (\mathcal{H}, \mathcal{C})
    \]

    \State \textbf{Solve Optimization Problem}
    \begin{align*}
    \argmin_{\{u_t\}_{t=\tc}^{\tc+\tp-dt}} & \sum_{t=\tc}^{\tc+\tp} \ell(x_t, u_t) + \ell_f(x_{\tc+\tp}) \\
    \text{subject to: } & \quad x_{t + dt} = f_{\theta_{t}}^{\RK}(x_t, u_t), \quad \forall t \in [\tc, \tc + \tp] \\
    & \quad x_t \in \mathcal{X}, \quad \forall t \in [\tc, \tc + \tp]  \\
    & \quad u_t \in \mathcal{U}, \quad \forall t \in [\tc, \tc + \tp - dt] 
    \end{align*}
    \State \textbf{Apply Control}
    \[
    u_{\tc} \gets u_{\tc}^*
    \]
    \State \textbf{Update History, State and Expected actions}
    \begin{align*}
    \mathcal{H} &\gets \mathcal{H} \cup \{ x_{\tc} \} \\
    x_{t} & \gets x_{t + dt} \\
    \mathcal{C} & \gets u_{t+dt}^* \forall t \in [\tc, \tc + \tp - dt]\\
    \tc & \gets \tc + dt
    \end{align*}
    \State \textbf{Repeat: Go to Step 1}
\end{algorithmic}
\end{algorithm}

\subsection{Pendulum experminet}

\subsubsection{Pendulum simulation}
\label{app:sim_setting}

% To setup the pendulum simulation, we used the settings presented in Table~\ref{tab:mujoco_sim_setting}.

\begin{figure}[b]
    \centering
    \includegraphics[width=0.95\linewidth]{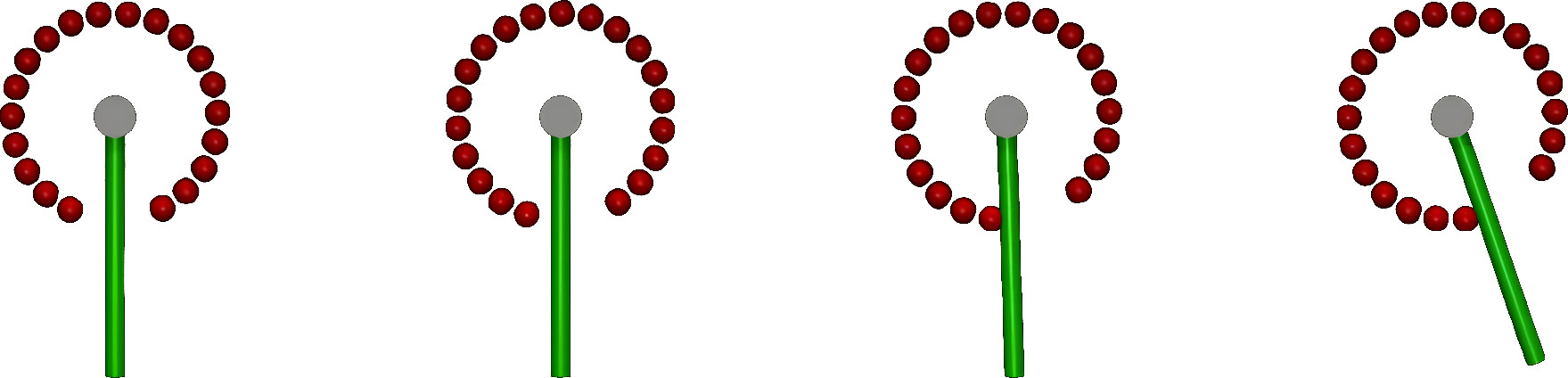}
    \caption{Visualization of the pendulum environment with backlash. Red dots are attached to the virtual joint, simulating the backlash and illustrating the slack between gears.}
    \vspace{-5mm}
    \label{fig:pendulum}
\end{figure}

We evaluated the proposed method in a pendulum environment similar to the one of the Gymnasium library~\cite{towers2024gymnasium} extended with backlash, a phenomenon commonly encountered in real-world robotics scenarios. The evaluation was carried out using a MuJoCo simulator~\cite{mujoco}, where the backlash was implemented as a joint limit, following the approach suggested in the user manual. To better visualize the backlash phenomenon, we added red dots attached to the virtual joint that simulates the backlash. The visualization of the environment is presented in Figure~\ref{fig:pendulum}. The pendulum system's dynamics model was derived from the Gymnasium implementation, with an added parameter to account for the gear ratio. The complete set of parameters describing the dynamics system includes mass, length, viscous friction, Coulomb friction, and gear ratio. The values of these parameters and the simulator settings are presented in Table~\ref{tab:mujoco_sim_setting}. Furthermore, implemented models were validated using the simulated pendulum without backlash to ensure that they could achieve zero training and validation error.

\begin{table}[h]
    \centering
    \caption{Pendulum simulation setting}
    \label{tab:mujoco_sim_setting}
    \begin{tabular}{|l|l|}
        \hline
        Parameter & Value \\
        \hline
        integrator & RK4 \\
        step size & $10^{-5}$s \\
        link mass & 1.0 \\
        link length & 0.5 \\
        link diameter & 0.02 \\
        damping & 0.05 \\
        frictionloss & 0.0 \\
        gear-ratio & 1.0 \\
        backlash & $30\deg$ \\        
        \hline
    \end{tabular}
\end{table}

We evaluated them on 20 episodes with initial position and velocities drawn from the range$[- \frac{\pi}{2}; \frac{\pi}{2}]\times[-0.5,0.5]$ range.

\subsubsection{Dataset}

To create the dataset, we defined a sampling procedure for the control signal (applied torque) and collected 360 ten-second episodes. These episodes were split into training, validation, and test datasets in a 70\%, 20\%, and 10\% ratio, respectively. The control signal was generated by sampling 15 random B-spline control points from a uniform distribution $[-2, 2]$, which were then interpolated. The resulting signal was clipped to the range $[-1, 1]$, making it impossible to swing up the pendulum directly. This sampling procedure enriched the dataset with saturated control inputs.

\subsubsection{Models training}

For training models utilizing longer sequence lengths, a prediction horizon of 250 samples was chosen, with a sampling rate of 100 Hz. All methods and models were trained for 2000 epochs, which was sufficient to observe convergence.
Before feeding the joint angle to the neural network, it was encoded using $sin(q)$ and $cos(q)$ functions.
% Neural network inputs were encoded using joint angle representations $sin(q)$ and $cos(q)$. 
Furthermore, prior to being fed into the neural network, the angular velocities were normalized by the maximum value of the training dataset.

\subsubsection{Pendulum MPC design}

The used cost function is the negated reward from the Gymnasium environment extended with terminal cost and control derivative penalization in terms of matrix R:
\begin{equation}
\begin{aligned}
    &x = [q, \dot{q}, \tau] \\
    &x_{ref} = [\pi, 0, 0] \\
    &u = \dot{\tau} \\
    \ell (x, u) = (&x - x_{ref})Q(x - x_{ref})^T + uRu^T\\
    \ell_f (x) = 10 \cdot (&x - x_{ref})Q(x - x_{ref})^T
\end{aligned}
\label{eq:pendulum_cost}
\end{equation}
where $Q$ is a diagonal matrix with weights 1, 0.1, 0.001 and $R$ set to $10e^{-8}$.
The state of the pendulum is defined by the angle $q$, where $q=\pi$ represents the upright position, the angular velocity $\dot{q}$, and the torque applied to the joint $\tau$. We follow a common MPC design practice, and we control the time derivative of the torque to smooth out the control actions.

\subsection{Drone}

\subsubsection{Drone simulation}

\begin{figure}[h]
  \centering
  \includegraphics[width=0.5\linewidth]{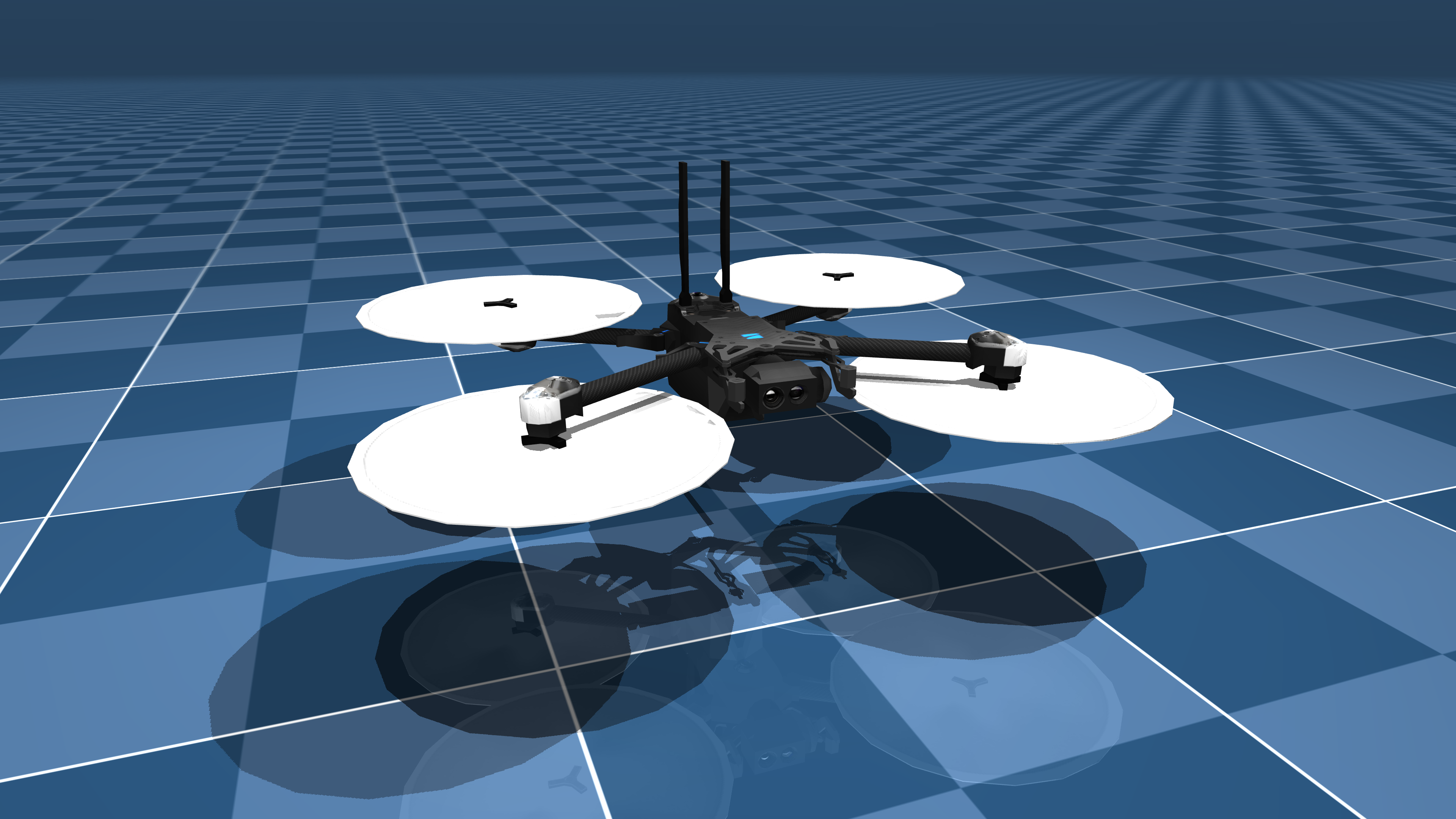}
  \caption{Base drone model {Skydio~X2} additionally extended with rope attached payload.}
  \label{fig:drone_sim}
  \vspace{-2mm}
\end{figure}

The drone environment was built in \textsc{MuJoCo} using the publicly
available \emph{Skydio~X2} model from the MuJoCo Menagerie
repository.\footnote{\url{https://github.com/google-deepmind/mujoco_menagerie/tree/main/skydio_x2}}
To introduce an unmodeled degree of freedom, we attach to the center of mass of the vehicle a massless tendon with a point mass payload placed at its end.

\subsubsection{Dataset}

Reference trajectories of orientation and altitude were randomly sampled and tracked with cascaded PID controller.
For each length of the rope, we recorded $300$ episodes of $20\,\mathrm{s}$ in $100\,\mathrm{Hz}$ and split the data $70\%/20\%/10\%$ into training, validation and test sets.

\subsubsection{Model training}

Similarly as for the the pendulum, all models were optimized with a multistep prediction loss computed over $100$‑step roll‑outs.
The nominal dynamics of the drone was represented with a rigid body dynamics as proposed in \cite{delayMPCdrone2020}.
% For the nominal dynamics model, we refer to the rigid body dynamics as in \cite{delayMPCdrone2020}.

\subsubsection{MPC formulation}

We formulate a trajectory tracking problem as in \cite{delayMPCdrone2020}, penalizing deviations from the reference position and attitude with quadratic weights $\operatorname{diag}(100,100,100,10,10,10,0.001,0.001,0.001,1,1,1)$ in
$\langle x,y,z,q_w,q_x,q_y,q_z,v_x,v_y,v_z, \omega_x, \omega_y, \omega_z \rangle$,
and a control cost of $0.01$ on individual motor thrusts.
To obtain a similar distribution of data in the training and downstream tasks, we used the test set as the reference trajectory on which the control performance is evaluated. The MPC used a prediction horizon of 1.6 s with a control frequency equal to 100Hz.

\subsubsection{Effect of rope length}

\begin{figure}[h]
  \centering
  \includegraphics[width=0.75\linewidth]{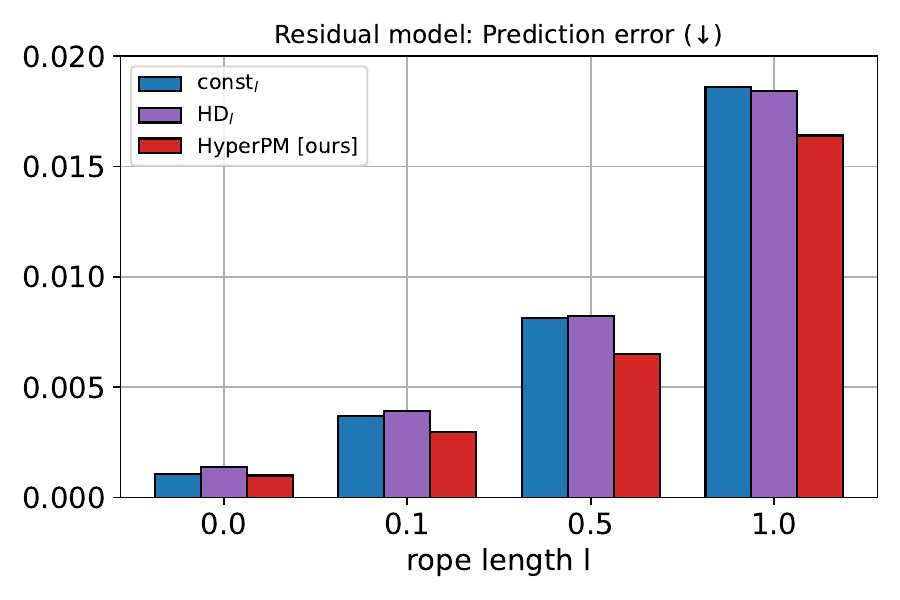}
  \vspace{-3mm}
  \caption{Comparison of prediction performance on datasets with different rope attachment lengths.}
  \label{fig:drone_rope_len}
  \vspace{-2mm}
\end{figure}

In this experiment, we evaluate the hypothesis that our method can anticipate the dynamics of unmodeled system states and partially compensate its effects on the observed states. 
To this end, we construct four datasets with progressively increasing rope lengths, thereby amplifying the influence of the unmodeled state---the swing of the rope-suspended payload. 
We employ a residual dynamics model to compensate for effects that can be inferred from the system's instantaneous state. 
Figure~\ref{fig:drone_rope_len} reports the predictive performance across all modeling approaches. 
When the rope length is zero---i.e., the payload is rigidly attached to the drone frame---our method performs comparably to a model with constant parameters. 
However, as the influence of the unmodeled state increases, the advantage of our method, $\HyperPM$, over the baselines, $\constl$ and $\HDl$, becomes evident. 
These results indicate that our method leverages time-varying parameters to recover hidden system states and predict their evolution, resulting in lower prediction errors.

% In this experiment we wanted to evaluate the hypothesis that our method can anticipate the dynamics of unmodeled states of the system.
% To do so, we created 4 datasets with increasing rope length, thereby increasing the influence of the unmodeled state, the swing of the rope suspended payload.
% We used residual dynamic model to compensate effects that can be derived from system instantaneous state.
% The Figure. \ref{fig:drone_rope_len} presents predictive performance for all modeling aproaches. 
% For case where the rope lenght is equal to 0 - payload is rigetly attachet to drone frame our method perform equalu well as model with constant parameters. But if the influence of unmodeled state is more pronounced the difference between our method $HyperPM$ and $\constl$ as well as $\HDl$ is evident.
% This suggests that our method can leverage time-variating parameters to recover hidden system state and predict their evolution resoulting in lower prediction errors.

% Due to simulation nature of this example we tested significant influence of unmodeled state - swing of rope suspended payload, we evaluated the predictive performance across different rope lengths presented on Figure. \ref{fig:drone_rope_len}.  A one can see on drone without unmodeled state, e.i. rope with 0 length all methods perform equally well as residual network as all effect can be described by use of drone state.
% With an increase in rope length, the difference is more pronounced.

\subsection{F1TENTH experiment}

\subsubsection{Experimental setup}
\label{sec:setup}
\begin{figure}[h]
    \centering
    \includegraphics[height=34.5mm]{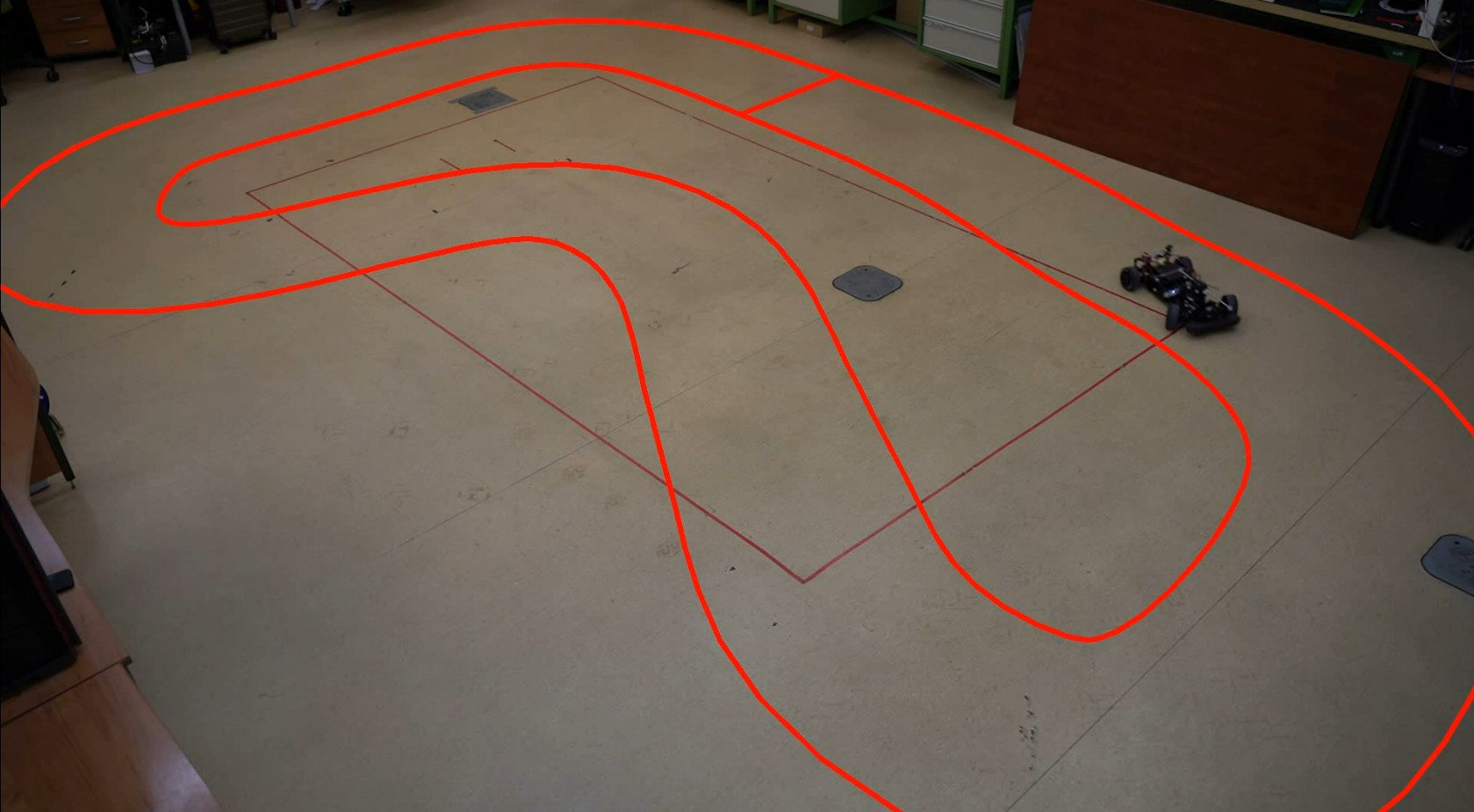}
    \includegraphics[height=34.5mm]{figures/autko.jpg}
    % \vspace{-0.9cm}
    \caption{The experimental setup for the F1TENTH racing task. The left image presents the virtual track in the laboratory, while the right one is the close-up of the vehicle we used.}
    \label{fig:setup}
\end{figure}

In this paper, we tackle the autonomous racing task using the real F1TENTH vehicle. The experimental vehicle, an Xray GTXE’22, is a 1/8th scale RC car with a four-wheel drive, powered by a single motor. To ensure a fair comparison between methods, we race in an empty space with a virtually defined racetrack. This ensures that the track is exactly the same for all methods and experiments, as it does not depend on the manual setting. Moreover, for accurate pose and velocity measurements during racing, we rely on the high-precision OptiTrack motion capture system. Our experimental setup is presented in Figure~\ref{fig:setup}.

\subsubsection{Analytical model}

For the purpose of car racing, we utilize the dynamics model of the single-track vehicle, as in \cite{Liniger2015OptimizationbasedAR, srinivasan_holistic_2021}, extended with the more accurate tire model MF6.1~\cite{besselink2010improved} referred to in the following section as an analytical model. Given such dynamics, the state vector is expressed as 
\begin{equation}
\label{eq:car_state}
    x = [v_x, v_y, r]^T,
\end{equation}
where $v_x, v_y$ denotes longitudinal and lateral velocities, and $r$ is the yaw rate. The control input $u = [\delta, \omega]^T$ is composed of the steering angle $\delta$ and the wheel speed scaled by the tire radius $\omega$.
We assume that some parameters are known; this includes mass $m$, gravity constant $g$, and wheel base $L$.

\subsubsection{Dataset}
To train and evaluate all considered methods, we collected a dataset using the physical setup introduced in Section~\ref{sec:setup}. The dataset was collected by an expert human driving on different tracks, with varying ranges of slip angle and slip ratio, trying to cover the entire envelope of the operation region of the vehicle state space. 
During manual racing, we captured relevant data of the internal state of the car, such as motor RPM, motor q-axis current, and steering angle, as well as the information about its position and orientation using OptiTrack. Moreover, we obtained the velocity in the vehicle frame by differentiating the position and orientation measurements using the Savitzky-Golay filter~\cite{savgol}.
All of these data were collected synchronously at 100Hz. In total, we recorded 36 minutes of driving. 
The recorded dataset was then split into disjoint training, validation, and test sets in the proportion 70/20/10\%.
To generate sequences, we applied a sliding window with a stride of 5 samples. Each window consisted of 135 samples, 50 of which were allocated to the observation window and 85 to the prediction horizon.

\subsubsection{F1TENTH MPC design}

In order to compare prediction models in a downstream application, we formulate the autonomous vehicle racing task by extending the primary system state \eqref{eq:car_state} by Frenet position, following \cite{verschueren_towards_2014, srinivasan_holistic_2021, novi_real-time_2020}, represented by distance along the center-line of the track $s$, distance perpendicular to the center-line $n$ and heading difference $\mu$, and control signals, resulting in 
\begin{equation}
    x = [s, n, \mu, v_x, v_y, r, \omega, \delta]^T.
\end{equation}
To capture the actuator dynamics absent in the primary model $u = [\omega, \delta]^T$, the control input was reformulated as $u = [\omega_{ref}, \delta_{ref}]^T$ and modeled as a first-order linear system.
To compensate for the measurement delay and actuator delay, we constrain adequate first control actions in such a way as to restrict the change of $u$, taking into account the recent history of applied actions.

The controller's task is to minimize a cost function that is divided into two components: $\ell(x,u) = \ell_s(x,u) + \ell_p(x,u)$.
The primary objective $\ell_p(x)$ is defined as the negative distance traveled along the centerline of the racetrack.
Additionally, the safety objective $\ell_s(x,u)$ is composed of three components: $\ell_s(x,u) = \ell_{track}(x) + \ell_{ctrl}(x, u) + \ell_{slip}(x)$. The track-bound cost $\ell_{track}(x)$ is formulated as a barrier function $p(h(x))$ to incorporate the constraint $h(x) \geq 0$ into the cost function, where $p(h(x)) = \log(1 + \exp(-\alpha h(x)))$ with $\alpha = 200$, following a similar approach to \cite{krinner_mpcc_2024}.
Since excessive slip may lead to unrecoverable situations, a cost penalizing large wheel slip and slip angles is included: $\ell_{slip}(x) = q_{\beta} \beta_{reg} + q_{slong} (\omega - v_x)^2$, where $\beta_{reg}$ is defined as in \cite{kabzan_learning-based_2019-1, srinivasan_holistic_2021}.
To emphasize the predictive behavior of the vehicle, a cost based on the smoothness of the control signal is defined as $\ell_{ctrl}(x,u) = q_u(\omega - \omega_{ref})^2 + q_u(\delta - \delta_{ref})^2$.

For a fair comparison, the MPC was tuned to $\constl$ model baseline and held constant for all models we considered, with $q_u=2.0$, $q_{slong}=0.1$, $q_{\beta}=1$. We compare the performance of the MPC with different dynamics models by using the same cost function as the MPC stage cost $\ell(x,u)$.
For the sake of safety and to evaluate the generalization abilities of the dynamics models, we set the horizon of MPC to 80 shooting nodes with frequency of 33.3Hz, which results in 2.67s horizon, while during training time 0.85 s (at 100Hz) horizon was used.

\subsection{Ablation studies}

\subsubsection{Impact of the expected future trajectories}

One of the key components of the proposed $\HyperPM$ is the exploitation of the expected future trajectories in the process of predicting the time-varying dynamics model parameters. Intuitively, future actions may have a significant impact on the behavior of the unmodeled dynamics that we aim to capture with the predicted parameters. In this experiment, we want to evaluate this hypothesis, by comparing the original $\HyperPM$ with the one that does not condition on the expected trajectory of actions ($\HyperPM$ - EAC). Moreover, for a better perspective we also report  the results obtained by the HyperDynamics apporach~\cite{xian_hyperdynamics_2021}, which not only does not condition on the future controls but also predicts only a single parameter for the entire MPC horizon (as opposed to $\HyperPM$ - EAC).

The results of this comparison are presented in Table~\ref{tab:ablation_eac}.
One can see that removing the dependence on the expected actions results in worse prediction performance for both the Pendulum and F1TENTH. However, the accuracy decrease is different for each system. In the case of Pendulum with backlash, we observe only a slight deterioration of $\HyperPM$ - EAC, when compared to $\HDl$. This suggests, that in this case the main source of the $\HyperPM$ success lies in the ability to predict the parameter's trajectories, but still knowing the planned future controls gives some advantage. 
This advantage is even more clear in the case of F1TENTH racing, in which we observe that without the expected future actions $\HyperPM$ is less accurate than the $\HDl$. This suggests that the ability of predicting time-varying parameters may be misused if there is not enough information about the expected evolution of the system. These results confirms the importance of conditioning the parameters prediction on the trajectory of expected actions.

\begin{table}[h]
\centering
\caption{Long-horizon prediction errors of $\HyperPM$ and $\HyperPM$ without conditioning on expected actions ($\HyperPM$ - EAC), related to the HyperDynamics $\HDl$.}
\begin{tabular}{lrrrr}
    & \multicolumn{2}{c}{Pendulum with backlash} & \multicolumn{2}{c}{F1TENTH racing} \\
    \toprule
    Model &    Error &  Deterioration [\%] &    Error &  Deterioration [\%]\\
\midrule
    $\HyperPM$ (ours) &  \textbf{0.447} &  -  & \textbf{0.0123}  & -\\
    $\HyperPM$ - EAC & 0.475 & 5.89 & 0.0147 & 16.32\\
    $\HDl$ & 0.626 & 28.59 & 0.0137 & 10.22\\
\bottomrule
\end{tabular}
\label{tab:ablation_eac}
\end{table}

\subsubsection{Impact of the $\sigma_{\text{robust}}$ hyperparameter}
\label{sec:sigma_ablation}

In the proposed $\HyperPM$ architecture we introduced a $\sigma_{\text{robust}}$ hyperparameter, which controls the amount of noise added to the representation of the expected future controls during training. The rationale behind this is to reduce the focus of the $\HyperPM$ on the true controls from the dataset that are observed during the training phase. Instead, we want to make $\HyperPM$ robust to some variability in the expected controls, which is expected during the deployment phase, when future controls are predicted with MPC.
In Table~\ref{tab:sigma-results} we present the prediction losses obtained for 3 different choices of the $\sigma_{\text{robust}}$.
As expected, the smaller the noise the more accurate the $\HyperPM$ on the training set. However, the best results on the validation set are obtained for a moderate value of $\sigma_{\text{robust}} = 0.2$.

\begin{table}[h]
    \centering
    \caption{Prediction loss in the F1TENTH task for different $\sigma_{\text{robust}}$ values}
    \label{tab:sigma-results}
    \begin{tabular}{ccc}
        \hline
        $\sigma_{\text{robust}}$ & Train & Validation \\ \hline
        0.5  & 0.0084228 & 0.01168  \\
        0.2  & 0.0070832 & \textbf{0.010384} \\
        0.05 & 0.0062105 & 0.010678 \\ \hline
    \end{tabular}
    \vspace{-5mm}
\end{table}

\subsection{Interpretability}
% \todo{probably needs rewriting and referencin in the main text}

% \todo{tutaj potrzeba opisu}
An important benefit of choosing a physically derived dynamics model structure is the ability to interpret variations in parameters, which can suggest physical phenomena worth including in the model.
For instance, the analytical dynamics model of a F1TENTH car, which we used in the previous experiments, lacks the capability to represent load transfer during braking and acceleration.
To evaluate whether $\HyperPM$ captures this unmodelled effect, we compare the load-transfer predicted by the HyperPM with the load-transfer calculated based on the physical model of this phenomenon.

Folliwing~\cite{balkill2018performanceVehicle}, we represent the normal force acting on the front wheel by
\begin{equation}
    F_{zf} = \frac{m g l_r}{L} - \frac{m h_{cg} a_x}{L},
\label{eq:front_tire_normal_force}
\end{equation}
where $l_r$ is the distance from the center of gravity to the rear axle, $m$ is the vehicle mass, $h_{cg}$ is the center of gravity 
height, $a_x$ is the longitudinal acceleration, and $L$ is the wheelbase.

In turn, the $\HyperPM$ considers only a first part of \eqref{eq:front_tire_normal_force}, i.e., $F_{zf} = \frac{m g l_r}{L}$. However, by predicting a time-varying $l_r$ parameter, it can effectively capture the physical effect of varying normal force $F_{zf}$, which is not considered by the single-track model.

\begin{figure}[h]
    \centering
    \includegraphics[width=0.7\linewidth]{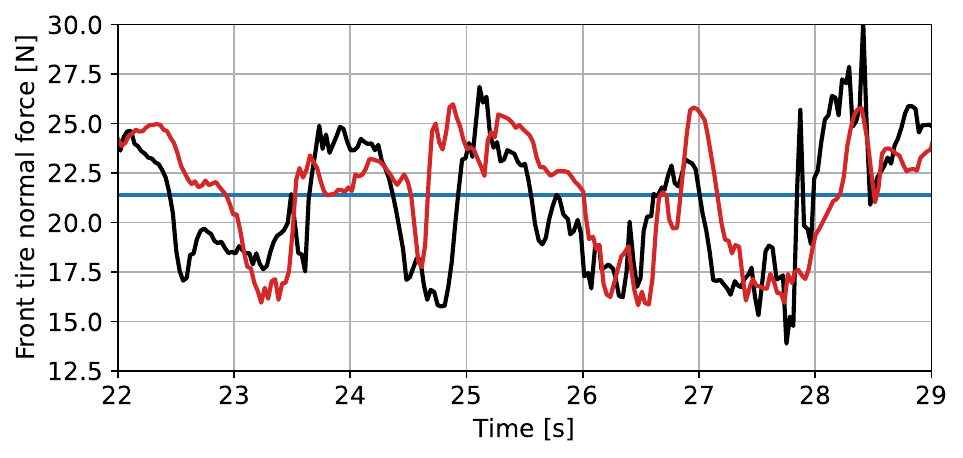}
    % \vspace{-3mm}
    \caption{Front tire normal force calculated using different methods. The blue line represents the front tire normal force calculated from the identified $\constl$ model. The black line shows the force calculated using~\eqref{eq:front_tire_normal_force}, while the red line represents the force obtained from $\HyperPM$.}
    \label{fig:front_tire_normal_force}
\end{figure}

In Figure~\ref{fig:front_tire_normal_force}, we compare the front tire normal forces $F_{zf}$ computed with~\eqref{eq:front_tire_normal_force} (black) and the ones predicted with $\HyperPM$ (red). A strong correlation between these predictions suggests that the $\HyperPM$ captures an unmodelled effect of the load transfer in an interpretable way, i.e. by virtually changing the position of the center of mass.

% To demonstrate that $\HyperPM$ can capture this effect, we calculated the front single-track tire normal force using prior physical knowledge as follows:
% \begin{equation}
%     F_{zf} = \frac{m g l_r}{L} - \frac{m h_{cg} a_x}{L}
% \label{eq:front_tire_normal_force}
% \end{equation}
% where  $l_r$ is the distance from the center of gravity to the rear axle, $m$ is the vehicle mass, $h_{cg}$ is the center of gravity 
% height, $a_x$ is the longitudinal acceleration, and $L$ is the wheelbase.

% As presented in Figure~\ref{fig:front_tire_normal_force} $\HyperPM$ predicts the normal force of the front tire as: $F_{zf} = \frac{m g l_r}{L}$
% where $l_r$ is a time-varying parameter that effectively captures the physical effect not modeled by the single-track model.

\subsection{Hyperparameters used in the experimental evaluation}
\label{app:hyperparameters}

The following tables summarize the hyperparameters used for all models in our experimental evaluation.  
Tables~\ref{tab:pendulum_hyperpm}--\ref{tab:pendulum_res} present the details of the pendulum swing-up, while Tables~\ref{tab:car_hyperpm}--\ref{tab:car_res} provide the corresponding values for the F1TENTH racing.  

All models were trained using the AdamW optimizer with a learning rate of $5\times10^{-4}$. 
For models trained on long sequences, the gradient clipping by value was set to $5\times10^{-4}$.
Additionally, all relevant models utilized a GRU-based time series encoder.

% \subsubsection{Pendulum}
% \FloatBarrier
\begin{table}[t]
    \centering
    \caption{Hyperparameters for $\HyperPM$  - Pendulum}
    \label{tab:pendulum_hyperpm}
    \begin{tabular}{|l|l|}
        \hline
        Hyperparameter & Value \\
        \hline
        B-spline control points & 25 \\
        GRU hidden size & 4 \\
        Causal MLP features per control point & 8 \\
        $\sigma_{robust}$ & 0.2 \\
        $n_{reg}$ & 0.05 \\
        $\Delta_{max}$ & 1.0 \\
        Batch size & 128 \\
        Activation & Tanh \\
        \hline
    \end{tabular}
\end{table}

\begin{table}[t]
    \centering
    \caption{Hyperparameters for $\HDl$ - Pendulum}
    \label{tab:pendulum_hdl}
    \begin{tabular}{|l|l|}
        \hline
        Hyperparameter & Value \\
        \hline
        GRU hidden size & 32 \\
        MLP size & 2 layers, 128 hidden size \\
        Batch size & 128 \\
        Activation & ReLU \\
        \hline
    \end{tabular}
\end{table}

\begin{table}[t]
    \centering
    \caption{Hyperparameters for $\HDs$ - Pendulum}
    \label{tab:pendulum_hds}
    \begin{tabular}{|l|l|}
        \hline
        Hyperparameter & Value \\
        \hline
        GRU hidden size & 4 \\
        MLP size & 2 layers, 512 hidden size \\
        Batch size & 256 \\
        Activation & ReLU \\
        \hline
    \end{tabular}
\end{table}

\begin{table}[t]
    \centering
    \caption{Hyperparameters for $\res$ - Pendulum}
    \label{tab:pendulum_res}
    \begin{tabular}{|l|l|}
        \hline
        Hyperparameter & Value \\
        \hline
        Residual NN size & Single hidden layer, 16 neurons \\
        Activation & Tanh \\
        \hline
    \end{tabular}
\end{table}

\begin{table}[t]
    \centering
    \caption{Hyperparameters for $\HyperPM$ residual as well as nominal dynamics - Drone}
    \label{tab:drone_hyperpm}
    \begin{tabular}{|l|l|}
        \hline
        Hyperparameter & Value \\
        \hline
        B-spline control points & 8 \\
        GRU hidden size & 16 \\
        Causal MLP features per control point & 8 \\
        $\sigma_{robust}$ & 0.05 \\
        $\Delta_{max}$ & 1.0 \\
        Batch size & 128 \\
        Activation & Tanh \\
        \hline
    \end{tabular}
\end{table}

\begin{table}[t]
    \centering
    \caption{Hyperparameters for $\HDl$ residual as well as nominal dynamics - Drone}
    \label{tab:drone_hdl}
    \begin{tabular}{|l|l|}
        \hline
        Hyperparameter & Value \\
        \hline
        GRU hidden size & 16 \\
        MLP size & 2 layers, 128 hidden size \\
        Batch size & 128 \\
        Activation & ReLU \\
        \hline
    \end{tabular}
\end{table}

\begin{table}[t]
    \centering
    \caption{Hyperparameters for Residual Model - Drone}
    \label{tab:drone_res}
    \begin{tabular}{|l|l|}
        \hline
        Hyperparameter & Value \\
        \hline
        Residual NN size & Single hidden layer, 32 neurons \\
        Activation & Tanh \\
        \hline
    \end{tabular}
\end{table}

\begin{table}[t]
    \centering
    \caption{Hyperparameters for $\HyperPM$ - F1TENTH}
    \label{tab:car_hyperpm}
    \begin{tabular}{|l|l|}
        \hline
        Hyperparameter & Value \\
        \hline
        B-spline control points & 8 \\
        GRU hidden size & 16 \\
        MLP features per control point & 32 \\
        $\sigma_{robust}$ & 0.5 \\
        $n_{reg}$ & 0.2 \\
        $\Delta_{max}$ & 0.5 \\
        Batch size & 128 \\
        \hline
    \end{tabular}
\end{table}

\begin{table}[t]
    \centering
    \caption{Hyperparameters for $\HDl$  - F1TENTH}
    \label{tab:car_hdl}
    \begin{tabular}{|l|l|}
        \hline
        Hyperparameter & Value \\
        \hline
        GRU hidden size & 16 \\
        MLP size & 2 layers, 128 hidden size \\
        Batch size & 256 \\
        \hline
    \end{tabular}
\end{table}

\begin{table}[t]
    \centering
    \caption{Hyperparameters for $\HDs$ - F1TENTH}
    \label{tab:car_hds}
    \begin{tabular}{|l|l|}
        \hline
        Hyperparameter & Value \\
        \hline
        GRU hidden size & 16 \\
        MLP size & 2 layers, 128 hidden size \\
        Batch size & 256 \\
        \hline
    \end{tabular}
\end{table}

\begin{table}[t]
    \centering
    \caption{Hyperparameters for $\res$ - F1TENTH}
    \label{tab:car_res}
    \begin{tabular}{|l|l|}
        \hline
        Hyperparameter & Value \\
        \hline
        Residual NN size & Single hidden layer, 64 neurons \\
        Activation & Tanh \\
        \hline
    \end{tabular}
\end{table}

\end{document}

%% file: definitions.tex
\DeclareMathOperator*{\argmin}{arg\,min}
\DeclareMathOperator{\HyperPM}{\text{HyperPM}}
\DeclareMathOperator{\RK}{RK4}
\DeclareMathOperator{\NN}{NN}

\newcommand{\x}{x}
\newcommand{\xpred}{\hat{\x}}

\newcommand{\xt}{\x(t)}
\newcommand{\xpredt}{\xpred(t)}

\newcommand{\dxpred}{\dot{\xpred}}

\newcommand{\tp}{t_p}
\newcommand{\tc}{t_c}
\newcommand{\thist}{t_o}

\newcommand{\consts}{\text{const}_{{s}}}
\newcommand{\constl}{\text{const}_{{l}}}
\newcommand{\HDs}{\text{HD}_{{s}}}
\newcommand{\HDl}{\text{HD}_{{l}}}
\newcommand{\res}{\text{res}}